\documentclass[10pt,twocolumn,letterpaper]{article}

\usepackage[pagenumbers]{iccv} 
\usepackage{multirow}

\definecolor{iccvblue}{rgb}{0.21,0.49,0.74}
\usepackage[pagebackref,breaklinks,colorlinks,allcolors=iccvblue]{hyperref}
\usepackage{utfsym}
\usepackage{wrapfig}

\def\paperID{13887} 
\def\confName{ICCV}
\def\confYear{2025}

\title{
\begin{center} 
\includegraphics[width=2.0cm,height=0.38cm]{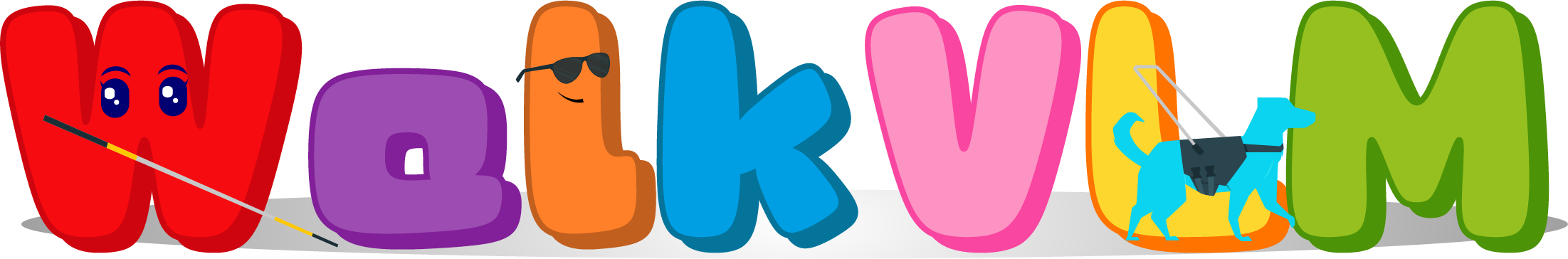} 
    : Aid Visually Impaired People Walking by Vision Language Model
\end{center}
}

\author{
\hspace{-5mm} Zhiqiang Yuan$^{\dagger}$, Ting Zhang$^{\dagger}$, Ying Deng, Jiapei Zhang, Yeshuang Zhu, Zexi Jia, Jie Zhou, Jinchao Zhang$^{\star}$ \\
Pattern Recognition Center, WeChat AI, Tencent Inc, China\\
}

\begin{document}
\maketitle
\begin{abstract}

Approximately 200 million individuals around the world suffer from varying degrees of visual impairment, making it crucial to leverage AI technology to offer walking assistance for these people.
With the recent progress of vision-language models (VLMs), applying VLMs to offer walking guidance has become popular. However, the existing methods of walking guidance are mainly based on self-curated question-answering datasets that are not publicly accessible, without a standardized benchmark for training or evaluation. 
Moreover, walking assistance often requires real-time streaming video analysis and the generation of concise yet informative reminders, making VLMs struggle due to excessive responses and low efficiency in inferences. 
In this paper, we introduce the first large-scale dataset dedicated to walking assistance, comprising 12,000 video-annotation pairs, 
to provide a unified benchmark for training and evaluating systems to help visually-impaired individuals walk. 
Furthermore, a WalkVLM model is proposed, which employs chain of thought for hierarchical planning to generate concise but informative reminders and utilizes temporal-aware adaptive prediction to reduce the temporal redundancy of reminders.
Finally, we have established a solid benchmark for blind walking task and verified the advantages of WalkVLM in stream video processing for this task compared to other VLMs.
Our dataset and code are available at the anonymous link \textcolor{blue}{\href{https://walkvlm2024.github.io}{https://walkvlm2024.github.io}}.

\end{abstract}    
\vspace{-10px}
\section{Introduction}
\label{sec:intro}

Approximately 200 million people worldwide suffer from varying degrees of visual impairment, with 36 million completely blind \cite{brady2013visual, real2019navigation}. 
These visually impaired people (VIPs) are facing severe challenges in daily activities such as walking, which may be alleviated by contemporary artificial intelligence technologies \cite{yang2024viassist, kuzdeuov2024chatgpt}.

The current walking assistance works are primarily based on electronic assistance devices, sensory substitution devices, and computer-vision-based assistance systems~\cite{giudice2020use, jain2023want, hersh2008assistive}.
Furthermore, vision-based assistance systems can be roughly categorized into detection-based methods and semantic-based methods~\cite{xia2023dataset, gurari2018vizwiz, yang2022seeway}.
Detection-based methods have been extensively studied, focusing on identifying potential obstacles within the field of view to help visually impaired persons (VIPs) walk safely~\cite{wang2024visiongpt, liu2023open}.
In contrast, semantic-based methods utilize vision-language models (VLMs) to analyze images and generate responses to the questions of VIPs~\cite{zhao2024vialm, xie2024emerging}.
With the recent advancements in VLMs~\cite{bordes2024introduction, du2022survey}, semantic-based methods have gained significant attention within the community.
Some studies have applied VLMs in zero-shot settings to evaluate their effectiveness in blind walking assistance~\cite{xie2024emerging, wang2024visiongpt}, while another line of work fine-tuned VLMs using traditional visual question-answer (VQA) datasets or small-scale and self-constructed datasets to improve the quality of the responses~\cite{yang2024viassist, merchant2024generating}.
These efforts have demonstrated the potential of VLMs in walking guidance tasks, yielding promising results and paving the way for the application in this area.

\begin{figure}[!t]
    \centering
    \includegraphics[width=0.9\linewidth]{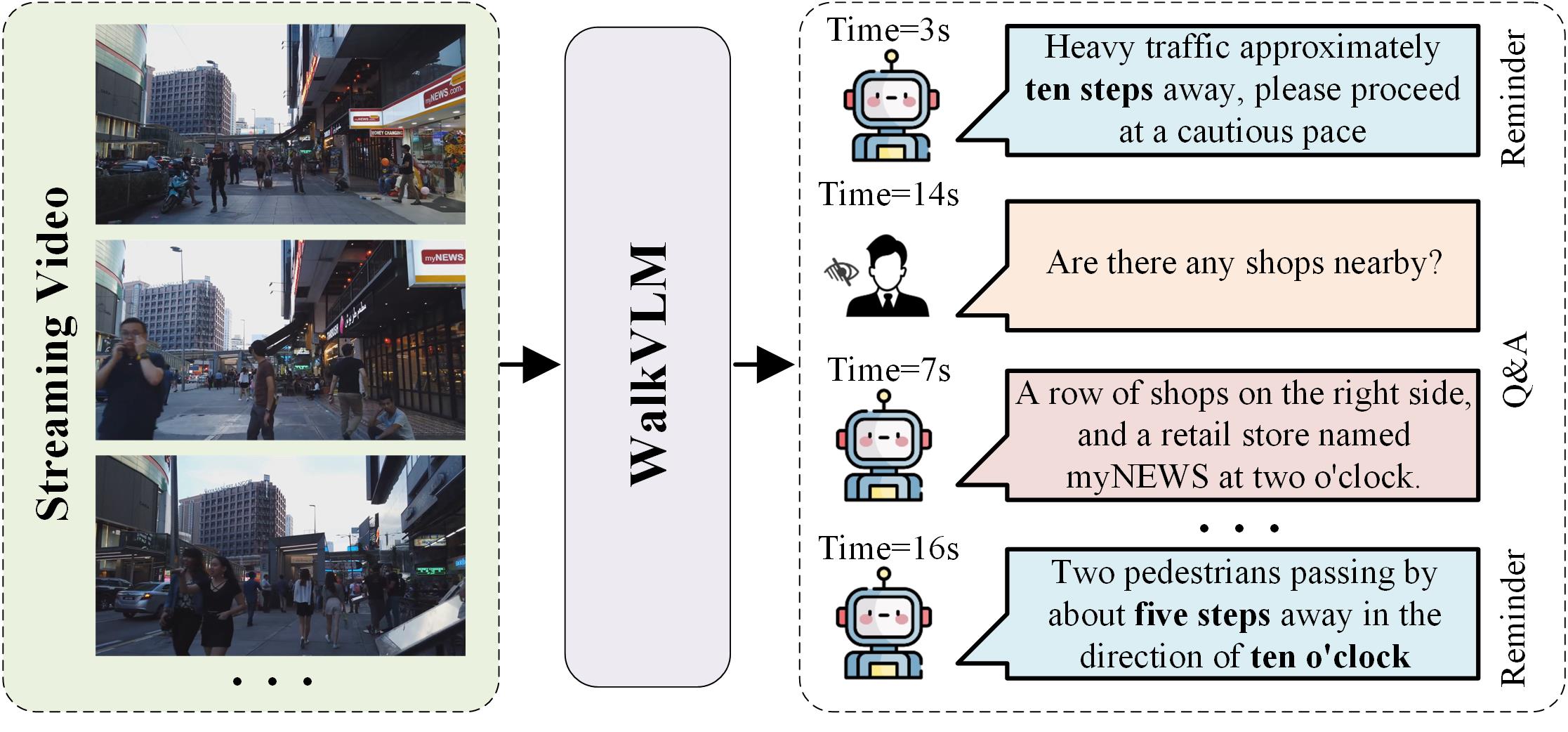}
    \vspace{-5pt}
    \caption{
WalkVLM delivers timely, concise, informative walking reminders for visually impaired individuals by utilizing hierarchical planning and temporal-aware adaptive prediction.
    }
    \label{fig1}
\vspace{-20pt}
\end{figure}

While VLM-based models for walking assistance have shown promise, several challenges remain as obstacles to effective application in real-world scenarios. First, most recent research~\cite{merchant2024generating, zhang2024vision} relies on limited image-text pairs, and there lacks a comprehensive and standardized benchmark. Additionally, the existing datasets~\cite{zhao2024vialm, de2024vqask} primarily adopt a question-and-answer paradigm, allowing VLMs to respond to explicit queries but making it challenging for VLMs to proactively generate context-aware guidance. Second, blind walking requires opportune streaming video analysis and the generation of concise yet informative reminders. However, existing approaches~\cite{arefeen2024vita, zhang2024sparsevlm} struggle with excessive redundancy in the generated responses and suffer from low efficiency in inference, posing significant challenges for practical applications.

To address these challenges, we propose WalkVLM, a vision-language model designed for blind walking assistance, and establish a new benchmark to facilitate the research in this field. Specifically, we introduce the Walking Awareness Dataset (WAD), a large-scale and diverse dataset comprising 12k annotated videos collected from Europe and Asia, providing a fair and standardized foundation for both training and evaluation of this task.
As illustrated in Figure~\ref{fig1}, WalkVLM processes streaming video inputs by incorporating a chain-of-thought reasoning framework to hierarchically guide the model in generating concise and informative reminders, and achieves opportune reminders by the proposed temporal-aware adaptive prediction. 
Comprehensive experiments demonstrate that WalkVLM generates more concise guidance and exhibits superior temporal adaptability when handling video streaming in blind walking task compared to existing VLM methods.


Our main contributions can be summarized as follows:
\begin{itemize}

    \item We construct a diverse and extensive dataset named the Walking Awareness Dataset (WAD), providing extensive data support for the blind walking task.

    \item We propose WalkVLM, a vision-language model for streaming video analysis, which adaptively generates concise yet informative walking guidance for visually impaired people.
    
    \item 
    To our knowledge, we are the first to utilize VLMs to provide opportune and reasoning walking guidance for visually impaired individuals, establishing a strong foundation for the practical application of VLMs in this field.
\end{itemize}
\vspace{-5px}
\section{Related Work}
\noindent
\textbf{Vision Datasets for Walking Guidance.}
Existing datasets for blind walking can be roughly divided into two types: detection-based~\cite{zhang2024grfb,islam2024dataset,xia2023dataset,tang2023dataset} and semantic-based~\cite{gurari2018vizwiz,zhao2024vialm}.
Detection-based datasets have been extensively studied in blind walking guidance, where researchers utilize these datasets to train models for obstacle detection, thereby reducing the accident rates of VIPs for this task.
For example, Zhang~$et\ al.$~\cite{zhang2024grfb} recently developed a TP-Dataset for detecting visual tactile paving surfaces and offered walking route guidance for the visually impaired.
Islam $et\ al.$~\cite{islam2024dataset} introduced a dataset containing 90 object annotations from 31 video clips to improve real-time object recognition systems to aid VIPs in navigation tasks.
Compared with detection-based datasets, semantic-based datasets are relatively limited, primarily containing data for question answering and providing an enhanced human-computer interaction experience.
Gurari~$et\ al.$~\cite{gurari2018vizwiz} constructed a visual question answering (VQA) dataset for VIPs containing 31k visual questions, each with 10 answers annotated by crowdsourcing.
In addition, there have been several self-built question-answering datasets with specific attributes~\cite{zhao2024vialm,yang2024viassist}, however, these datasets are not publicly available and are relatively small in scale, making them unsuitable for large-scale and unified benchmarking.

\begin{figure}[!t]
    \centering
    \includegraphics[width=1.0\linewidth]{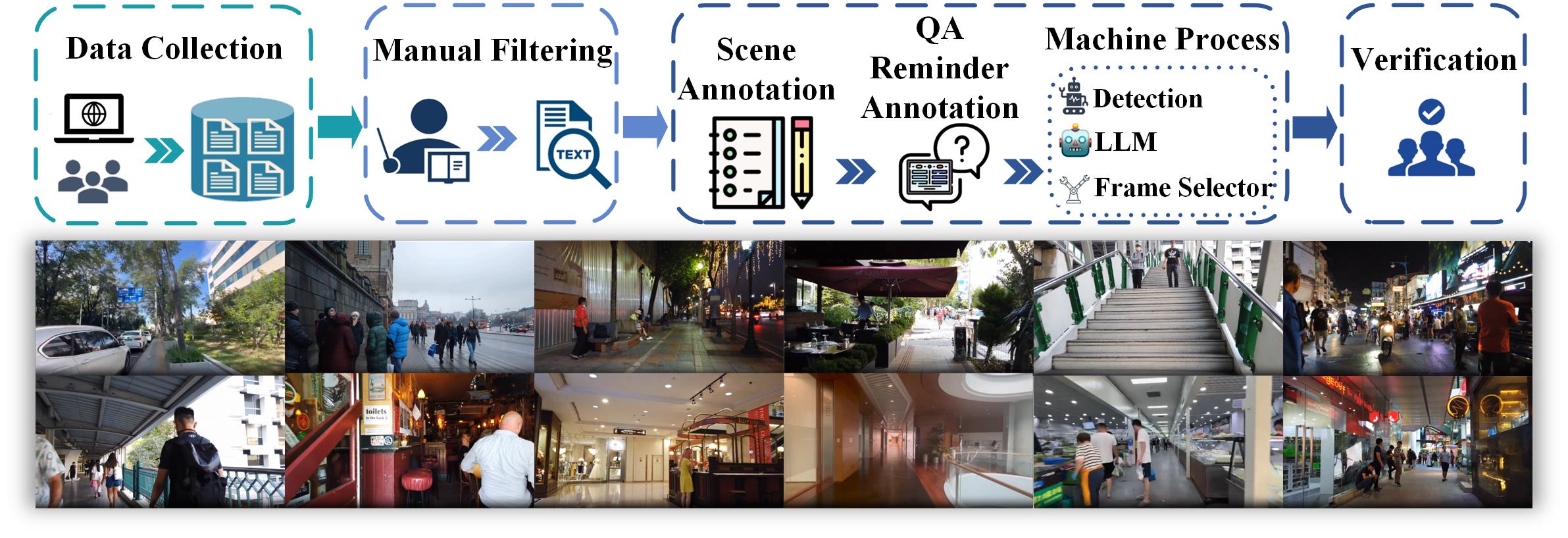}
    \vspace{-20pt}
    \caption{
    The data annotation pipeline for constructing the Walking Awareness Dataset (WAD). 
    Appendix \textcolor{red}{A.5} provides more examples to illustrate the diversity of WAD dataset.
    }
    \label{fig:label_process}
    \vspace{-15pt}
\end{figure}

\noindent
\textbf{Vision-based Methods for Walking Guidance.}
Similar to the categorization of datasets, the vision-based methods for walking guidance can also be divided into detection-based methods~\cite{wang2024visiongpt, liu2023open} and semantic-based methods~\cite{zhao2024vialm}.
Detection-based methods typically use detectors to identify obstacles during walking and provide users with specific object locations.
Liu~$et\ al.$~\cite{liu2023open} proposed a system for understanding open scenes, which improves detection performance by using SAM~\cite{kirillov2023segment} to generate pixel-level dense segmentation masks.
Tian~$et\ al.$~\cite{tian2021dynamic} introduced a method to understand dynamic crosswalk scenes, including crosswalks, vehicles, and pedestrians, to provide VIPs with guidance of when and where to cross the road.
In contrast, the semantic-based approach provides VIPs with the scene understanding in the form of question-answering.
Merchant~$et\ al.$~\cite{merchant2024generating} verified that vision-language models can generate correct and informative instructions for VIPs, and studied methods to provide users with context-related guidance.
Yang~$et\ al.$~\cite{yang2024viassist} explored how to utilize VLMs to provide reliable visual question answers for VIPs, and fine-tuned the VLMs with LoRA on a small amount of self-constructed dataset to generate detailed and practical suggestions.
Moreover, a few applications such as Be My AI \footnote{https://www.bemyeyes.com} have adopted semantic-based methods to enable VIPs to take photos for visual question answering.
However, these applications only support the question-and-answer paradigm and struggle to provide concise and timely guidance during walking.

\noindent
\textbf{Vision-Language Models.}
With the advancements of large-language models (LMM), vision-language models have also gained significant attention~\cite{zhang2024vision, xing2024survey, jin2024efficient}.
Liu $et\ al.$ \cite{liu2023llava} introduced LLaVa, which integrates a ViT visual encoder to process images, followed by an MLP to map them to the LLM, achieving favorable performance in benchmark evaluations for vision-based question-answering tasks.
Building on LLaVa, numerous studies have emerged, expanding its applications across diverse fields~\cite{zhu2024llava, chen2024llava, li2024llava, shi2024math}. Additionally, multimodal models such as Qwen, Gemini, and MiniCPM-V~\cite{QwenVL, reid2024gemini, yao2024minicpm} have introduced support for multiframe image inputs and undergone optimizations for scenarios such as edge-device deployment, significantly improving the capacity of VLMs in a wide range of applications. Despite these advancements that demonstrated the potential of the large-scale multimodal models~\cite{tian2024drivevlm}, the applications in domain-specific tasks remain limited. For example, only a limited number of studies~\cite{merchant2024generating, yang2024viassist, zhao2024vialm} have explored the usage of VLMs for blind walking assistance, and a unified, systematic modeling approach for this task is still lacking.
\vspace{-5px}
\section{Walking Awareness Dataset}



\subsection{Data Collection}

The dataset encompasses diverse geographical sources collected from 10 different locations across Europe and Asia. Approximately 20\% of the dataset is derived from the recordings made by annotators, while the remainder of the data is collected from YouTube\footnote{https://www.youtube.com/@poptravelorg}. To ensure data variability, six recorders positioned their cameras at chest level, using focal lengths of 13mm, 20mm, and 26mm, with resolutions ranging from 1080p to 4K at 60fps. We compiled approximately 13 hours of walking videos in total and see Appendix \textcolor{red}{A} for the details on the duration of data collected from various regions.

\subsection{Annotation Strategy}

Figure \ref{fig:label_process} shows the overall annotation pipeline of the walking awareness dataset.
In the following sections, we detail the dataset construction process from two perspectives: scene annotation and response annotation.

\begin{table*}[!t]
\centering
\resizebox{0.95\linewidth}{!}{
\begin{tabular}{l|c|ccccccccc}
\toprule
                      Dataset & Type &  \#Sample    & Modality                                 & Bounding Box & Weather & Danger level & Scene Summary & QA & Reminder       & Open \\ \hline
      \rowcolor{gray!12} \multicolumn{1}{l|}{Obstacle Dataset (2023)\cite{tang2023dataset}} & $\mathcal{T}$ & 8k & Image & \checkmark            &       \usym{2717}             &      \usym{2717}        &           \usym{2717}    &  \usym{2717}    &         \usym{2717}          & \checkmark    \\
                         WOTR (2023)\cite{xia2023dataset} & $\mathcal{T}$ & 13k & Image                                 & \checkmark            &        \usym{2717}            &         \usym{2717}     &   \usym{2717}            &  \usym{2717}    &          \usym{2717}        & \checkmark    \\
                    \rowcolor{gray!12}     ISLAM $et\ al.$(2024)\cite{islam2024identifying} & $\mathcal{T}$ & 31 & Image \& Video                          & \checkmark            &         \usym{2717}           &       \usym{2717}       &      \usym{2717}         &   \usym{2717}   &     \usym{2717}         & \checkmark    \\
                                  Wang $et\ al.$(2024)\cite{wang2024visiongpt} & $\mathcal{T}$   & 50 & Video                        & \checkmark            & \checkmark                  &      \usym{2717}        &          \usym{2717}     &    \usym{2717}  &   \usym{2717}           &   \usym{2717}   \\
                                 \hline
  \rowcolor{gray!12} VizWiz (2018)\cite{gurari2018vizwiz}  & $\mathcal{S}$ & 31k & Image                             &    \usym{2717}          &          \usym{2717}          &       \usym{2717}       &      \usym{2717}         & \checkmark    &         \usym{2717}         & \checkmark    \\
                                  Zain $et\ al.$(2024)\cite{merchant2024generating} & $\mathcal{S}$    & 48 & Image                        &        \usym{2717}      &              \usym{2717}      &       \usym{2717}       &          \usym{2717}     &   \usym{2717}   & \checkmark                 &    \usym{2717}  \\ \hline
                            \rowcolor{gray!12}      WAD (Ours)    &  $\mathcal{T}$$\mathcal{S}$   &    12k / 120k & Video / Image                                 & \checkmark            & \checkmark                  & \checkmark            & \checkmark             & \checkmark    & \checkmark                & \checkmark    \\ \bottomrule
\end{tabular}
}
\vspace{-5pt}
\caption{
Static information comparison of different datasets in blind walking assistance task.
$\mathcal{T}$ and $\mathcal{S}$ denote the dataset types, representing target-based and semantic-based datasets, respectively.
WAD dataset holds a significant advantage in terms of sample numbers, categories, and modalities.
}
\label{dataset_comp}
\vspace{-15pt}
\end{table*}

\noindent
\textbf{Scene annotation.}
Scene annotation aims to label the inherent attributes of the recorded scenes.
We requested nine annotators to label each video scene in terms of weather conditions, location type, traffic flow rating, danger level, and scene description.
Weather conditions are categorized into six types (\eg, sunny, rainy) for outdoor scenes, while indoor scenes are left unclassified.
Location types are divided into eight categories, such as corridors and pedestrian walkways.
The traffic flow rating is divided into three levels, determined based on the number of visible pedestrians in the video stream.
The danger level reflects potential walking hazards in the scenes, which are qualitatively assessed by the traffic flow rating and road surface smoothness. 
Scene description provides a high-level summary of the environment, incorporating details such as pedestrian and vehicle movement, road conditions, and surrounding context.
Subsequently, we employed the open-world detection model \cite{zhou2022detecting} for the preliminary object detection, 
followed by manual verification to ensure the data reliability.

\begin{figure}[!t]
    \centering
    \includegraphics[width=0.75\linewidth]{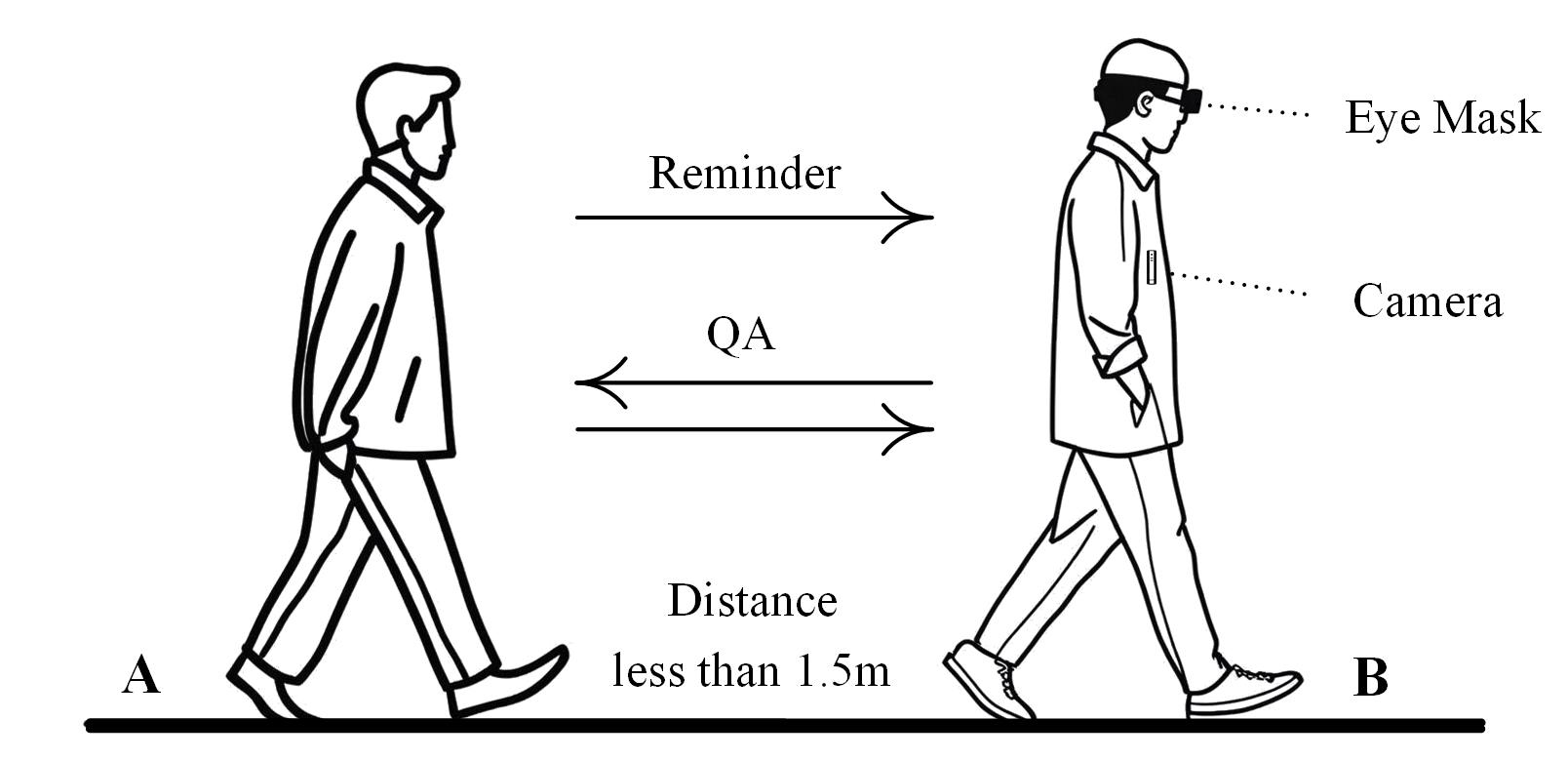}
    \vspace{-7pt}
    \caption{
    Blind test experiment for analyzing the most critical information needed by users in blind walking.
    We required two individuals to collaborate as a team, where the participant at the rear provided directions to enable the individual at the front to arrive at a specific location safely in the absence of any visual information.
    }
    \label{fig:rule_from}
    \vspace{-10pt}
\end{figure}

\noindent
\textbf{Response annotation.}
The response in our dataset refers to the concise instructions that the model aims to generate, as well as the answers that directly address the user's inquiries in the task of blind walking assistance. To identify the most critical information needed by users in this context, we conducted a blind test experiment, as illustrated in Figure~\ref{fig:rule_from}. In this experiment, two participants collaborated in pairs: Person A, positioned behind, provided verbal directions, while Person B, wearing an eye mask, followed these instructions to reach a designated destination without collisions. This setup ensures that Person B relies solely on the verbal instructions from Person A, eliminating any pre-existing route knowledge that a VIP might possess and enabling us to analyze what information is necessary for this task. Through extensive trials, our findings confirm that the provided guidance enables visually impaired individuals to navigate safely, indicating that the information provided by Person A is sufficiently effective for Person B. To further refine our analysis, we recorded both video and audio of the experiments, examined the communication between participants, and subsequently identified key types of information essential for the blind walking guidance task. These types of information serve as the basis for subsequent \textit{reminder} and \textit{QA} annotations.


\begin{figure}[!b]
    \centering
    \includegraphics[width=0.9\linewidth]{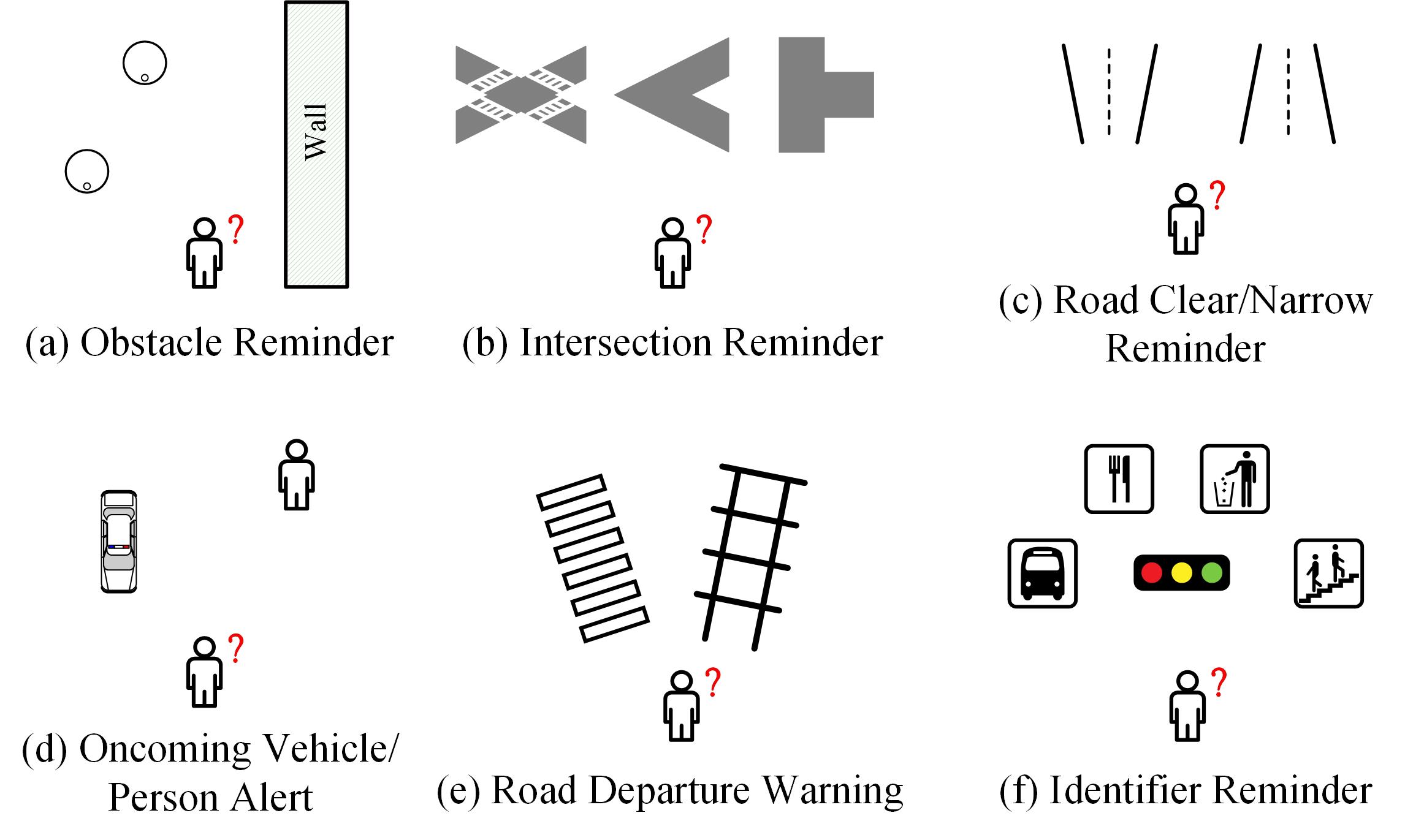}
    \vspace{-7pt}
    \caption{
Visualization of six scenarios that require reminders, which were summarized through multiple blind experiments.
    }
    \label{fig:sixcase}
    \vspace{-15pt}
\end{figure}

\begin{itemize}
    \item \textit{Reminder type.}
Based on the blind test experiment, as shown in Figure \ref{fig:sixcase}, we categorize the reminders during walking into six types.
(a) Obstacle reminder: Trigger a reminder when there is a non-moving obstacle on the walking route.
(b) Intersection reminder: Trigger a reminder when the current road has intersections, turns, $etc$.
(c) Road clear/narrow reminder: Provide reminders about the width and passability of the road.
(d) Oncoming vehicle/person reminder: When there are moving obstacles on the walking route, trigger a reminder for potential dangers.
(e) Road departure warning: Issue a warning when there is an angular offset between the walking route and the current road.
(f) Identifier reminder: Provide reminders for prominent landmarks in the scene, such as road signs and traffic lights.

    \item \textit{QA type.}
For QA type, we summarize based on three aspects: scene perception, road inquiry, and detailed consultation.
(a) Scene perception: Macro-level insights such as the understanding of the scene.
(b) Road inquiry: Route planning to reach a certain location within visible range.
(c) Detailed consultation: Knowledge QA on local details, such as road sign content, shop names, $etc$.

\end{itemize}

When annotating reminder and QA, we asked nine annotators to indicate the specific location of obstacles in the video.
During annotating, the distances are represented by steps on a scale of 5, and the directions are indicated by clock positions to reduce the offset caused by the camera perspective.
The annotation interface is shown in Appendix \textcolor{red}{A.3}.
After completing the annotations, we apply GPT \cite{dubey2024llama} to rephrase the annotated content and conduct manual verification to further standardize the annotation content to remove potential bias.

\subsection{Dataset Analysis}

\begin{figure}[!t]
    \centering
    \includegraphics[width=1.0\linewidth]{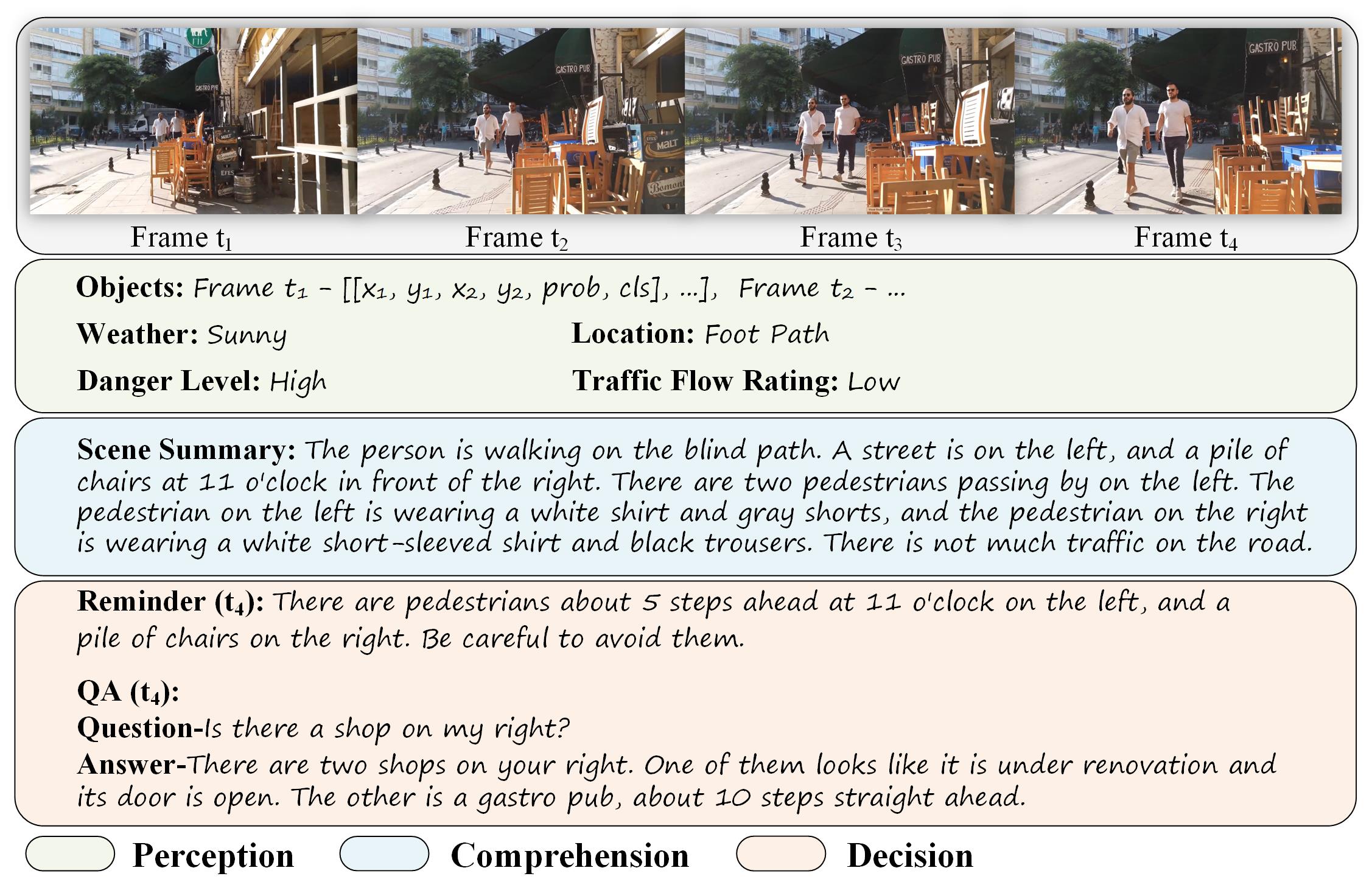}
    \vspace{-20pt}
    \caption{
    Visualization of the Walking Awareness Dataset.
    Each sample contains a video clip and multiple annotations, with the hierarchy divided into perception, comprehension, and decision.
    }
    \label{fig:data_sample}
    \vspace{-10pt}
\end{figure}

Figure~\ref{fig:data_sample} shows an example of the WAD dataset, and we divide the annotations into three groups according to lower to higher semantic levels: perception, comprehension, and decision.
The perception label reflects the basic attributes, such as obstacle location, weather conditions, $etc.$, while the comprehension label indicates the model's understanding of the entire scene.
The decision contains reminder and QA, reflecting the model's decision on the user's walking based on its understanding of the current scenario.

\begin{figure}[!t]
    \centering
    \includegraphics[width=1.0\linewidth]{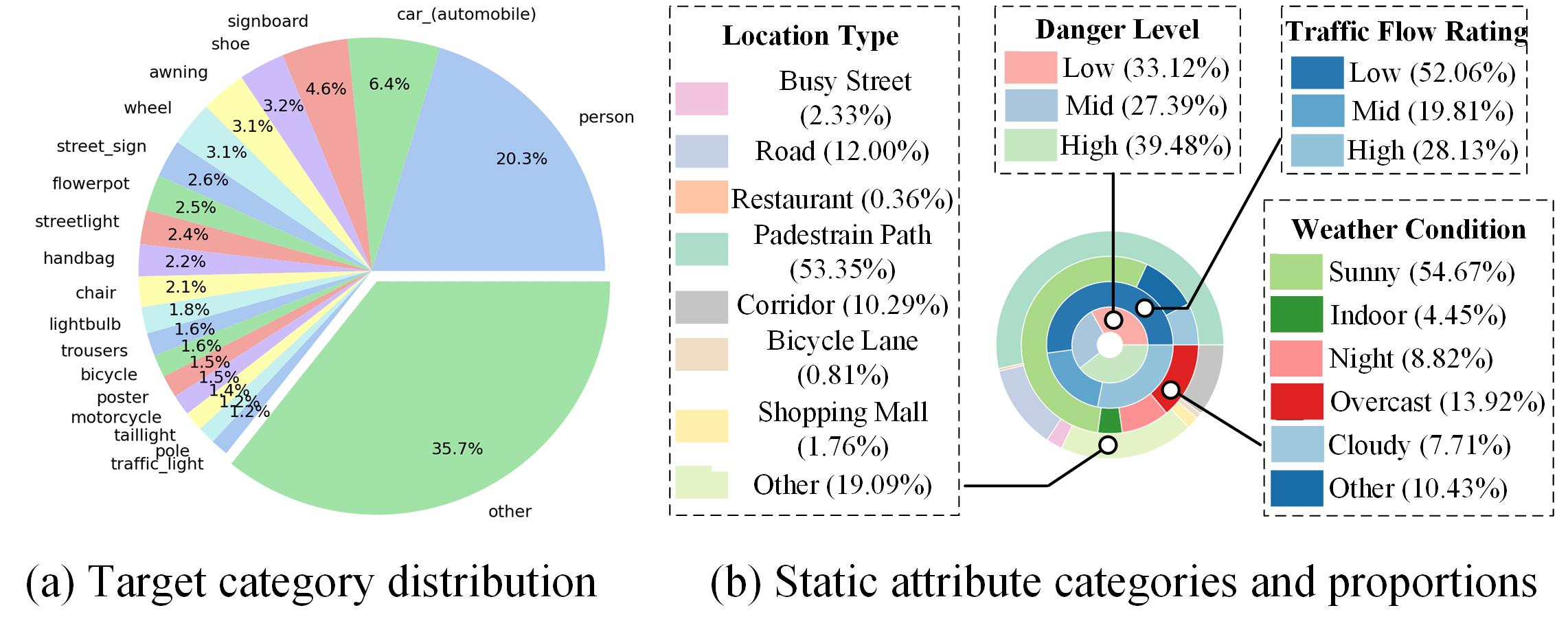}
    \vspace{-20pt}
    \caption{
    Visualization of the proportion of targets and categories in our Walking Awareness Dataset.
    }
    \vspace{-18pt}
    \label{fig:count_static}
\end{figure}

Table~\ref{dataset_comp} illustrates the comparison between the WAD dataset and other prevalent datasets utilized in blind walking tasks, with $\mathcal{T}$ representing the detection-based dataset and $\mathcal{S}$ indicating the semantic-based dataset.
Compared to other datasets, WAD includes more extensive data while containing more static attributes of the environment, scene summaries, QA, and reminders, thus providing more supervision to train the model.
It is worth emphasizing that the samples we provide are exclusively video clips, which possess a greater volume of information in comparison to the images supplied by other datasets. Moreover, for each video clip, we extracted 10 keyframes to streamline the usage in broader research.
The walking awareness dataset contains 3.47 million instances, and we show categories and the respective proportions in Figure \ref{fig:count_static}(a).
The distribution of categories in the WAD dataset is shown in Figure~\ref{fig:count_static}(b).
We selected 1.5k samples as a test set based on different static tag types, reminder types, and QA types to ensure diversity and completeness in evaluation.

\begin{figure*}[!t]
    \centering
    \includegraphics[width=1.0\linewidth]{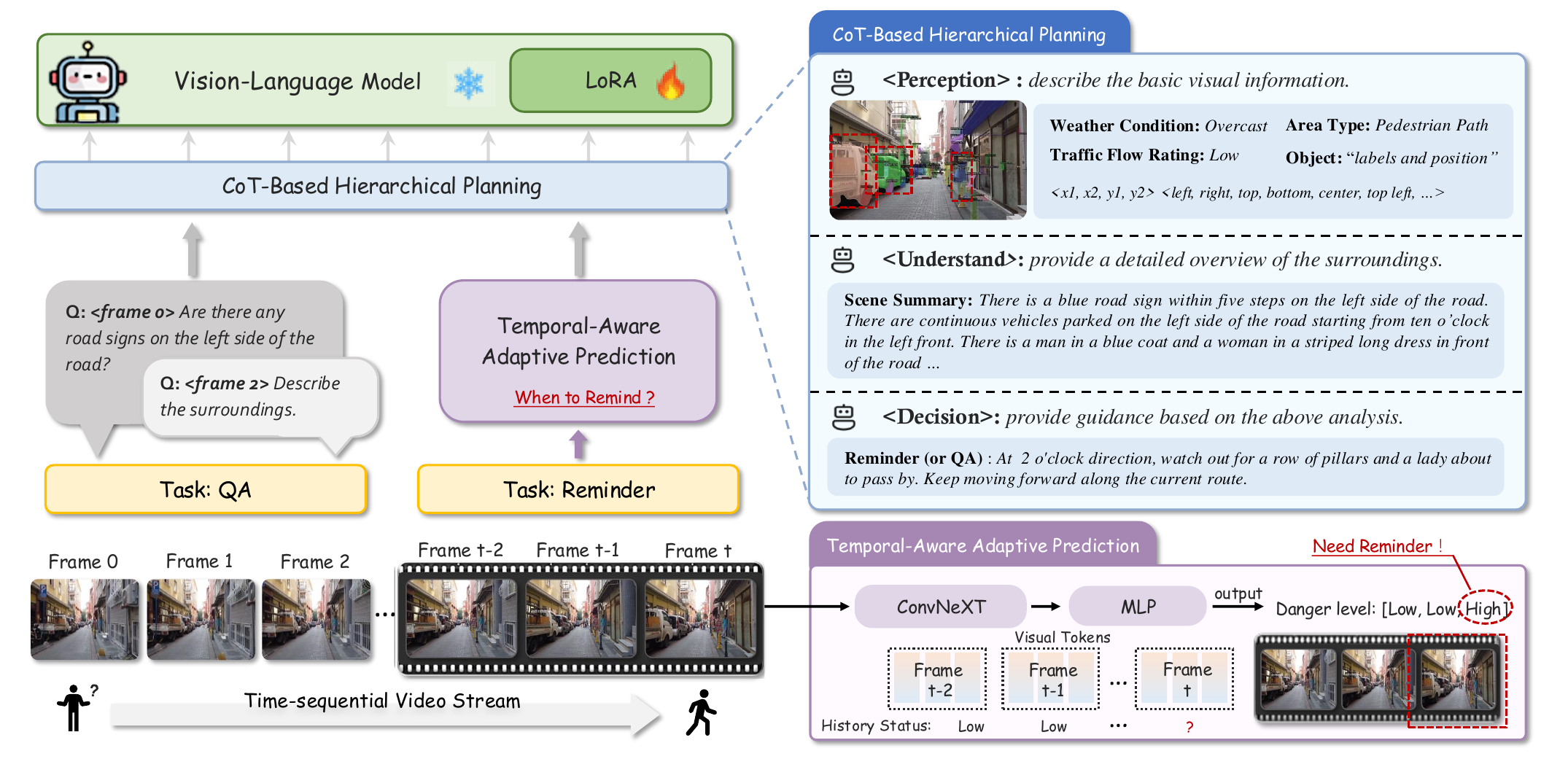}
    \vspace{-17pt}
    \caption{
    An overview of the proposed WalkVLM framework. 
    WalkVLM employs CoT-based hierarchical planning to summarize the static attributes and understanding of scenes, thereby facilitating the subsequent reminder generation and QA tasks.
    Furthermore, temporal-aware adaptive prediction is proposed to calculate the trigger state of VLM, thereby reducing the temporal redundancy of outputs.
    }
    \label{fig:main_cot}
    \vspace{-13pt}
\end{figure*}


\vspace{-3mm}
\subsection{Further Explanation of Data}



The WAD dataset is utilized for model training to generate auxiliary information that assists edge groups in making walking decisions. While guide dogs and canes are commonly used by VIPs in real-world scenarios, the WAD dataset primarily focuses on capturing the most critical visual information. Future iterations of the dataset will incorporate additional elements to further enhance the applicability. Given the limited availability of participants within the VIP community, we adopt a blind test experiment to determine appropriate annotation types. To assess the alignment of the obtained annotation types with the needs of VIPs, we conducted 34 validated surveys in which VIPs evaluated the applicability of the annotation types to walking. 
\begin{table}[!h]
\resizebox{\linewidth}{!}{
\begin{tabular}{l|>{\columncolor{gray!12}}cc>{\columncolor{gray!12}}cc>{\columncolor{gray!12}}c}
\hline
      & Positive & Relatively Positive & Neutral & Relatively Negative & Negative \\ \hline
Ratio & 41.18\%  & 47.06\%             & 8.82\%  & 2.94\%              & 0.00\%   \\ \hline
\end{tabular}
}
\vspace{-10px}
\caption{
The attitude proportion of the VIP group towards the annotation types of the WAD dataset.
}
\vspace{-8px}
\label{blv}
\end{table}
As shown in Table~\ref{blv}, approximately 88.24\% of the participants expressed a positive attitude towards the WAD dataset, especially appreciating the reminders at traffic intersections. Respondents with a neutral or lower attitude indicated the potential redundancy in the annotations and expressed concerns about the risk of inaccurate prompts from the model.


%

\vspace{-8px}
\section{WalkVLM}
\vspace{-4px}

The overall architecture of WalkVLM is shown in Figure~\ref{fig:main_cot}.
We start with problem formulation and proceed with the introductions of the hierarchical planning and temporal-aware adaptive prediction to generate concise and opportune walking reminders.

\subsection{Problem Formulation}
\label{Formulation}
We aim to guide a VLM to process video streams, enabling it to provide walking reminders with temporal attributes and respond to specific queries in human-machine interactions.
Specifically, at time $t_0$, given a sequence of the newly appeared frames $[ I_{t_{-N}} ..., I_{t_{-1}}, I_{t_{0}} ] $,
and the corresponding categlory and obstacle positions in the image $[ O_{t_{-N}} ..., O_{t_{-1}}, O_{t_{0}} ]$, VLM is expected to generate a concise and informative reminder $T^R_{t_0}$ based on visual context.
While walking, VIPs can pose a question $Q_{t_0}$ to communicate with the VLM at any time to request information regarding the current scene and route.
However, generating reminders at every frame may degrade the walking guidance experience and impose significant real-time processing pressure on the hardware. To address this, WalkVLM predicts the current VLM trigger state $s_{t_{0}}$ based on historical states $[ s_{t_{-N}} ..., s_{t_{-1}}, s_{t_{0}} ] $ and the previous $N$ frames, to determine optimal moments for issuing reminders.

\vspace{-5px}
\subsection{CoT-Based Hierarchical Planning}
\label{Planning}

To enhance the reasoning capabilities of the VLM, we use Chain of Thought (CoT) \cite{wei2022chain} to enable the model to derive step-by-step conclusions by summarizing comprehensive information, including static attributes and scene summary. This approach refines raw visual inputs into concise and contextually relevant reminders. 
The model architecture integrates a vision transformer encoder with a large language model (LLM). 
The vision encoder generates image tokens, while an attention-based extractor aligns these tokens with the LLM, facilitating comprehensive information processing and understanding. 
WalkVLM aggregates multi-frame information to ensure a holistic perception of the surroundings, thereby improving the accuracy and reliability of walking guidance.

We structure the reminder generation process into three levels: perception, comprehension, and decision.
At the \textbf{perception level}, the model extracts static visual attributes from the current frame, such as location type, weather conditions, and traffic flow rating.
To enhance the VLM's ability to focus on critical elements and improve visual perception accuracy, we incorporate the Priori-Object Location Module (POLM). 
The POLM initially uses a generic object detector~\cite{zhou2022detecting} to identify and locate objects in the scene. It then filters these objects based on size and confidence scores to prioritize key elements that reflect road conditions and potential hazards. 
The filtered information, along with fundamental environmental attributes, provides the model with sufficient representation of the external world. 
At the \textbf{comprehension level}, the model integrates all outputs from the perception stage, merging local detection results and fragmented scene information into a comprehensive global summary. 
Utilizing the capabilities of the VLM and the detailed attributes extracted in the perception stage, this phase ensures that the model attains a clear understanding of the current environment.
At the \textbf{decision level}, we focus on training the WalkVLM model to perform visual QA and generate reminders.
At this stage, the model possesses an understanding of the static attributes and overall context of the environment. 
With appropriate guidance, the model is expected to analyze potential hazards in the scene.

During training, we adopt a CoT approach to progressively feed information from three levels into the VLM, and during testing, the model predicts the aforementioned attributes and generates the corresponding responses.

\vspace{-5px}
\subsection{Temporal-Aware Adaptive Prediction}
\label{AdaptivePrediction}
Despite VLMs are capable of scene parsing across multiple frames and generating the required output, directly applying them to video streaming introduces several challenges.
First, generating walking reminders frame by frame or at fixed intervals can result in significant temporal redundancy, which can degrade the user experience.
Additionally, continuous VLM inference imposes computational pressure on hardware devices, making real-time processing inefficient. 
Addressing these challenges is crucial for the effective deployment of VLMs in video streaming processing.

To address the aforementioned issues, we introduce the Temporal-aware Adaptive Prediction (TAP) module that incorporates historical information to pre-determine whether the VLM should be triggered at a given moment, thereby alleviating the inference pressure on hardware.
Specifically, as shown on the right of Figure~\ref{fig:main_cot}, TAP utilizes a lightweight model to analyze historical $N$ frames and determine whether to trigger the VLM at the current moment based on the prior output states.
Given a sequence of frames $ [ I_{t_{-N}}, ..., I_{t_{-1}}, I_{t_{0}} ]$, a 3D convolutional model extracts the features $f_v$.
Simultaneously, the predicted trigger states from the previous $N$ moments are independently embedded, concatenated, and then passed through multiple layers of perceptrons to generate the state feature $f_s$.
The final trigger probability $\mathcal{P}_t$ is obtained by integrating $f_v$ and $f_s$ through an additional MLP.
To optimize reminder generation, TAP defines three trigger levels, corresponding to different degrees of environmental risk as categorized in the WAD dataset.
Subsequent experiments demonstrate that TAP effectively reduces temporal redundancy by providing proper triggers for the VLM, ensuring that VLM-based walking guidance remains both efficient and effective.

\begin{table*}[!t]
\centering
\resizebox{0.95\linewidth}{!}{
\begin{tabular}{l|ccccc|ccccc}
\hline
\multirow{2}{*}{Model} & \multicolumn{5}{c|}{Reminder Task}                                  & \multicolumn{5}{c}{QA Task}                                      \\ \cline{2-11} 
                       & TF-IDF & ROUGE-1 & ROUGE-2 & ROUGE-L & GPT Score & TF-IDF & ROUGE-1 & ROUGE-2 & ROUGE-L & GPT Score \\ \hline
\rowcolor{gray!12} LLaVa (7B)   \cite{liu2024visual}       & 0.061             & 0.062   & 0.005   & 0.070   &      0.013     & 0.074             & 0.084   & 0.011   & 0.072   &     0.012      \\
 DeepSeek-VL (1.3B)   \cite{lu2024deepseek}       & 0.073             & 0.098   & 0.015   & 0.090   &      0.148     & 0.182             & 0.103   & 0.020   & 0.095   &     0.336      \\
\rowcolor{gray!12} DeepSeek-VL (7B)     \cite{lu2024deepseek}        & 0.132             & 0.073   & 0.009   & 0.068   &       0.015    & 0.189             & 0.088   & 0.021   & 0.081   &       1.000    \\
Yi-VL (6B)   \cite{young2024yi}             & 0.112             & 0.093   & 0.009   & 0.085   &       0.133 & 0.113             & 0.091   & 0.012   & 0.082   &      0.168     \\ 
\rowcolor{gray!12} MiniCPM-V2.6 (8B)   \cite{yao2024minicpm}       & 0.111             & 0.071   & 0.007   & 0.064   &     0.025      & 0.192             & 0.139   & 0.025   & 0.120   &     0.832      \\
Qwen2-VL (7B) \cite{Qwen2VL}          & 0.106             & 0.107   & 0.010   & 0.097   &        0.044   & 0.232             & 0.182   & 0.037   & 0.162   &     0.504      \\ 
\rowcolor{gray!12} GPT-4o      \cite{hurst2024gpt}            & 0.116             & 0.078   & 0.008   & 0.072   &       -    & \underline{0.242}             & 0.163   & 0.034   & 0.145   &      -   \\ \hline
 \textcolor{black}{*DeepSeek-VL (7B)}   \cite{lu2024deepseek}       &  \textcolor{black}{0.129}             & \textcolor{black}{0.152}   & \textcolor{black}{0.043}   & \textcolor{black}{0.141}  &    \textcolor{black}{0.969}       & \textcolor{black}{0.193}             & \textcolor{black}{0.166}   & \textcolor{black}{0.037}   & \textcolor{black}{0.149}   &     \textcolor{black}{2.496}      \\ 
\rowcolor{gray!12} \textcolor{black}{*MiniCPM-V2.6 (8B)}   \cite{yao2024minicpm}         & \underline{\textcolor{black}{0.152}}             & \underline{\textcolor{black}{0.171}}   & \underline{\textcolor{black}{0.056}}   & \underline{\textcolor{black}{0.170}}    &      \underline{\textcolor{black}{1.024}}     & \textcolor{black}{0.207}             & \textcolor{black}{0.176}   & \textcolor{black}{0.044}   &  \textcolor{black}{0.160}   &      \textcolor{black}{3.172}      \\
 \textcolor{black}{*Qwen2-VL (7B)} \cite{Qwen2VL}           & \textcolor{black}{0.147}             &  \textcolor{black}{0.163}  & \textcolor{black}{0.054}   & \textcolor{black}{0.165}   &      \textcolor{black}{1.018}     & \textbf{\textcolor{black}{0.246}}             & \underline{\textcolor{black}{0.196}}   &  \underline{\textcolor{black}{0.047}}   & \underline{0.167}   &     \underline{3.246}     \\ \hline
\rowcolor{gray!12} WalkVLM                & \textbf{0.166}             & \textbf{0.191}   & \textbf{0.062}   & \textbf{0.173}   &   \textbf{1.103}        & 0.189 &	\textbf{0.202} &	\textbf{0.051} &	\textbf{0.174}       &    \textbf{4.168}       \\ \hline
\end{tabular}
}
\caption{Quantitative comparison of different methods on reminder generation and QA tasks. 
WalkVLM leads in almost all the TF-IDF, ROUGE, and GPT Score metrics.
* indicates the fine-tuned model.
\textbf{Bold} and \underline{underline} indicate the best and the second-best, respectively.
}
\label{quantitative}
\vspace{-13pt}
\end{table*}

\vspace{-9px}
\section{Experiments}
\vspace{-5px}

\subsection{Settings}

\textbf{Models \& Details.}
WalkVLM is implemented with the MiniCPM-V2.6 model \cite{yao2024minicpm}, which is an 8B multimodal model built upon Qwen2-7B \cite{yang2024qwen2}.
We add LoRA to all the linear layers of MiniCPM-V2.6 with a rank of 64 while maintaining the video stream sampling rate of 2 FPS.
The number of historical frames $N$ is set to 3, and the visual extraction backbone in the TAP module is ConvNext3D~\cite{liu2022convnet}.
We compared WalkVLM with multiple popular multimodal models, including GPT-4o~\cite{hurst2024gpt}, Qwen2-VL (7B)~\cite{Qwen2VL}, MiniCPM-V2.6 (8B)~\cite{yao2024minicpm}, DeepSeek-VL (1.3B\&7B)~\cite{lu2024deepseek}, Yi-VL (6B)~\cite{young2024yi}, LLaVa (7B)~\cite{liu2024visual}.
For open-source models with the same size, we use the WAD dataset to fine-tune them for comparison.
All the prompts for the large model usage can be found in the Appendix \textcolor{red}{B}.

\noindent
\textbf{Metrics.}
We use the following metrics to evaluate the models: 
\textbf{(a) ROUGE.}
This metric measures the similarity between the generated and the reference text by comparing overlapping words or phrases, including ROUGE-1, ROUGE-2, and ROUGE-L~\cite{lin2004rouge}.
\textbf{(b) TF-IDF Similarity.}
By combining term frequency and inverse document frequency to evaluate the weight of words, the text can be represented as a TF-IDF vector, and then the semantic similarity between texts is evaluated with TF-IDF similarity~\cite{ramos2003using}.
\textbf{(c) GPT Score.}
GPT4 is used to evaluate the superiority ratio between the generation results of different multimodal models and the ground truth (GT)~\cite{zheng2023judging, achiam2023gpt}.
\textbf{(d) Temporal Redundancy F1-Score (TRF).}
Given the historical model state and historical frames, the model predicts the danger level of the current moment, and the F1-Score is calculated between the prediction and the GT.
The evaluation metrics capture different aspects of performance: 
TF-IDF and ROUGE evaluate semantic similarity and word-level granularity, respectively, while the GPT Score determines the optimal result by comparing model outputs with GT.
Additionally, TRF quantifies the temporal redundancy of the model's output, where a higher TRF indicates lower redundancy in generated reminders.

\vspace{-5px}
\subsection{Quantitative Results}
Table~\ref{quantitative} presents the quantitative metrics of different models on the reminder and QA task.
On the ROUGE metric, WalkVLM achieved the highest scores in both tasks, confirming that the model's output is closest to the GT at the word granularity.
For semantic similarity, measured by TF-IDF, WalkVLM performs best in reminder tasks, indicating that its outputs are more concise and accurate in alignment with GT.
While in QA tasks, WalkVLM's performance on TF-IDF scores does not stand out significantly. 
This could be due to the training process, where the model is optimized to generate concise answers, potentially reducing its capacity to offer detailed explanations of the questions.
The GPT score represents the overall evaluation of the LLM on the generated results and the GT.
WalkVLM outperforms other models, including GPT-4o, in GPT scores for both reminder and QA tasks, demonstrating that the model’s outputs exhibit the most consistent distribution with the GT.

\begin{table}[!t]
\resizebox{\linewidth}{!}{
\begin{tabular}{l|>{\columncolor{gray!12}}cc>{\columncolor{gray!12}}cc>{\columncolor{gray!12}}c}
\hline
Model  & Yi-VL  & MiniCPM-V2.6  & GPT-4o & Qwen2-VL & WalkVLM \\ \hline
TRF  &     0.341 &   0.396                  &        0.430    &   \underline{0.449}    &            \textbf{0.505}
\\ \hline
\end{tabular}
}
\caption{
Temporal redundancy assessment of the reminder task, our method achieved the highest TRF score.
}
\label{trf}
\vspace{-12pt}
\end{table}

We use TRF to evaluate the temporal redundancy of the output of various VLMs.
Specifically, we provide the model with multiple frames of images along with historical dangerous states as inputs, allowing it to predict the current danger level and determine whether a reminder should be triggered.
As shown in Table~\ref{trf}, WalkVLM achieved the highest TRF score among all models, indicating its superior ability to reduce reminder redundancy over time.

\begin{figure}[!t]
    \centering
    \includegraphics[width=1.0\linewidth]{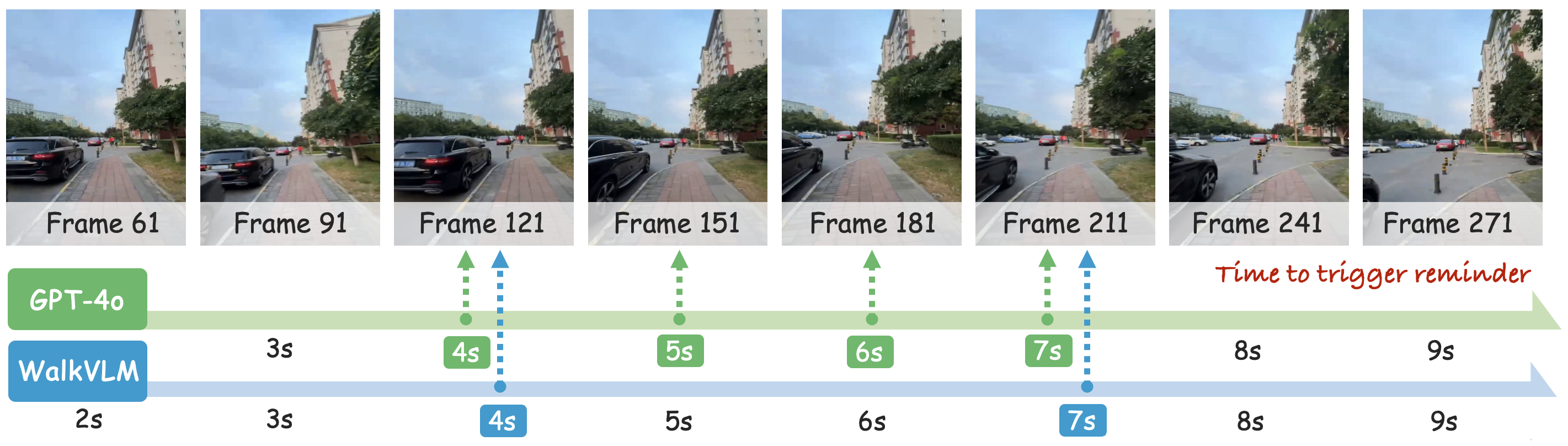}
    \caption{
    Visualization of triggering moments of GPT-4o and WalkVLM.
    WalkVLM triggers with less redundancy, providing information to users in a more timely manner.
    }
    \label{fig:reminder}
    \vspace{-13pt}
\end{figure}

\begin{figure*}[!t]
    \centering
    \includegraphics[width=0.9\linewidth]{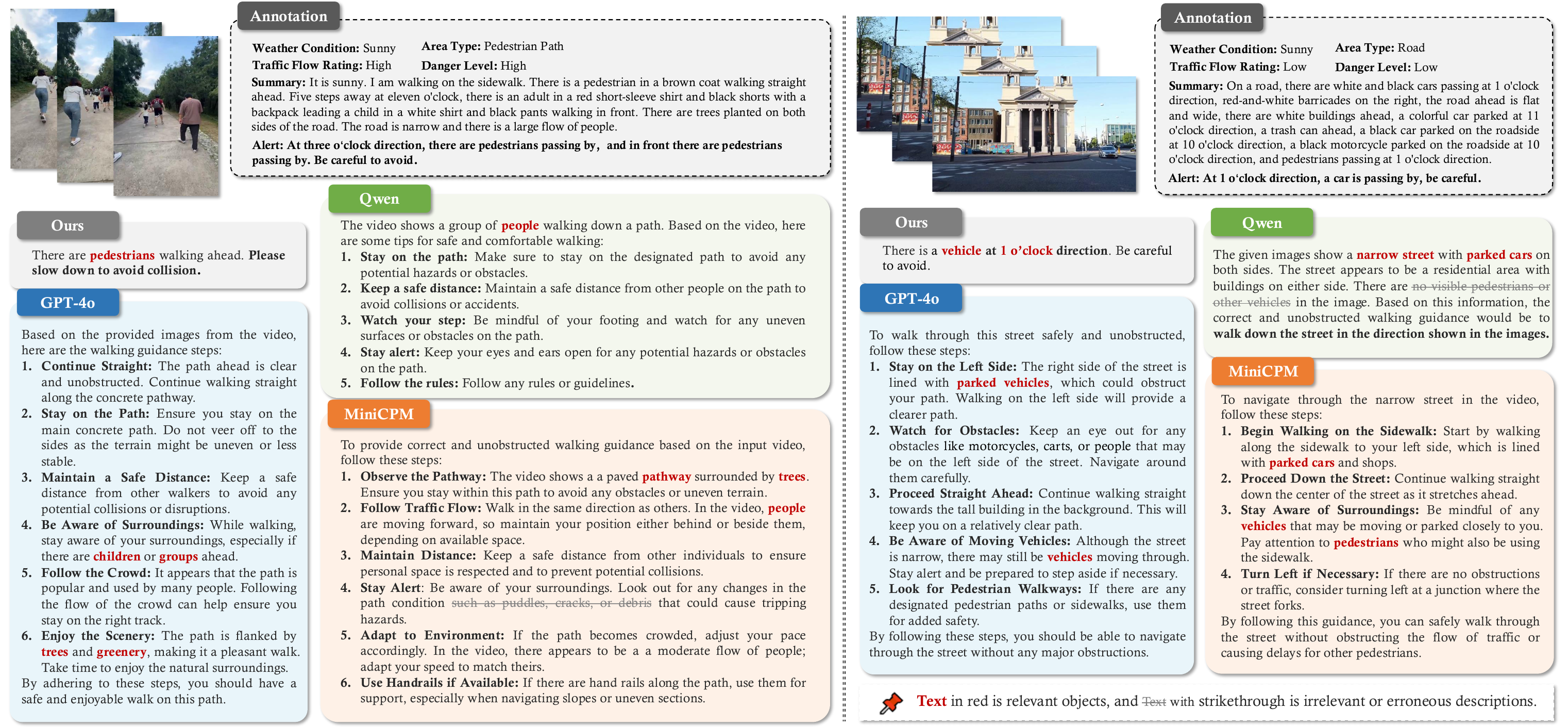}
    \caption{
    Visualization comparison of different VLM models.
    Compared to other models, WalkVLM is able to generate concise and informative answers, providing users with a good experience in blind walking.
    Refer to Appendix \textcolor{red}{C} for more visualization comparisons.
    }
    \label{fig:visualization_comparison}
    \vspace{-13pt}
\end{figure*}

\vspace{-5pt}
\subsection{Qualitative Results}
\vspace{-5pt}

Figure~\ref{fig:visualization_comparison} presents a visual comparison of the reminder task across different VLM models.
Compared to other methods such as GPT-4o, WalkVLM generates more concise and informative responses, enhancing the user experience.
In the left example, while other models provide highly detailed responses, WalkVLM delivers a concise prompt to the user, effectively emphasizing the most critical information.
In the right example, WalkVLM successfully identifies a car approaching from the one o’clock direction and conveys this essential detail to the user, while other models fail to achieve.
Figure~\ref{fig:reminder} further illustrates a qualitative comparison of GPT-4o and WalkVLM in terms of temporal redundancy. 
Our model triggers the VLM less frequently, reducing temporal redundancy while ensuring the timely delivery of crucial information.
Appendix \textcolor{red}{C} presents more qualitative results, including the comparison with other VLMs on real-world video streams.

\vspace{-5pt}
\subsection{Subjective Results}
\vspace{-5pt}

\noindent
\textbf{User Study.}
In Table~\ref{userstudy}, we recruited eight annotators to perform a subjective evaluation of various VLMs with respect to language conciseness and semantic similarity to the GT.
Participants were required to rank the results independently, and we measured the model performance using the top-1 superiority ratio, where a higher score indicates better performance.
\begin{table}[]
\centering
\resizebox{\linewidth}{!}{
\begin{tabular}{l|cc|cc}
\hline
\multirow{2}{*}{Model} & \multicolumn{2}{c|}{Reminder Task} & \multicolumn{2}{c}{QA Task} \\ \cline{2-5} 
                       & Concise.    & Semantic.    & Concise.  & Semantic.  \\ \hline
\rowcolor{gray!12} Yi-VL                 &      0.085       &  0.023            &     \underline{0.121}      &       0.022     \\ 
 GPT-4o                  &         0.056    &   0.195           &      0.030     &     \textbf{0.205}       \\ 
\rowcolor{gray!12} DeepSeek-VL(1.3B)           &     0.026        &   0.080           &     0.091      &        0.107    \\ \hline
*DeepSeek-VL(7B)             & 0.002            &   \underline{0.197}           &     0.044      &     0.114       \\
\rowcolor{gray!12} *MiniCPM-V2.6           &     0.026        &           0.122   &       0.061     &       0.190     \\
*Qwen2-VL            &      \underline{0.121}       &       0.168       &    0.077       &     \underline{0.192}       \\ \hline
\rowcolor{gray!12} WalkVLM                &      \textbf{0.683}       &       \textbf{0.216}       &       \textbf{0.576}    &      0.170      \\ \hline
\end{tabular}
}
\caption{
User study results on conciseness and semantic similarity across different tasks. Higher score indicates better performance. 
}
\label{userstudy}
\vspace{-16pt}
\end{table}
Compared with other methods, WalkVLM demonstrates a substantial improvement in terms of conciseness, both in reminder and QA tasks.
In terms of the semantic similarity to the GT, WalkVLM slightly outperforms GPT-4o in the reminder task but performs marginally worse in the QA task, which be attributed to its concise output style, which may occasionally omit details necessary to fully capture the semantics of the GT in QA tasks.

\noindent
\textbf{Closed-loop Analysis.}
We calculate the performance of WalkVLM when it is actually used for the VIPs.
\begin{wraptable}{r}{0.20\textwidth} 
\begin{tabular}{c>{\columncolor{gray!12}}cc>{\columncolor{gray!12}}c}
\hline
 & \footnotesize{p1} & \footnotesize{p2} & \footnotesize{p3} \\
\hline
 \footnotesize{w/} & \footnotesize{24} & \footnotesize{22} & \footnotesize{28} \\
 \footnotesize{w/o} & \footnotesize{35} & \footnotesize{33} & \footnotesize{37} \\
\hline
\end{tabular}
\vspace{-2px}
\caption{Comparison of the number of manual reminders required by VIPs during walking when using and not using WalkVLM.}
\captionsetup{font={small},skip=2pt, justification=raggedright}
\label{tab:actuallyexp}
\end{wraptable}
Table~\ref{tab:actuallyexp} presents the number of manual reminders required for three VIPs with varying degrees of visual impairment while traveling from point A to point B, and the comparative experiment evaluates the performance with and without WalkVLM, with experiments conducted first with WalkVLM to eliminate potential route memory bias.
In the experiment, the number of manual reminders required for the group with walkvlm was significantly less than that of the other experimental group, which fully demonstrates that our model can effectively support blind walking.

\vspace{-7pt}
\subsection{Ablative Study}
\vspace{-5pt}

\begin{table}[!t]
\centering
\resizebox{0.95\linewidth}{!}{
\begin{tabular}{l|cccc}
\hline
Configuration    & TF-IDF & ROUGE-1 & ROUGE-2 & ROUGE-L \\ \hline
\rowcolor{gray!12} w/o CHP          & 0.094  & 0.073   & 0.007   & 0.066   \\
w/o Pos Prior    & 0.151  & \underline{0.189}   & \textbf{0.062}   & \underline{0.171}   \\
\rowcolor{gray!12} w/o POLM Prior & \underline{0.152}  & 0.178   & 
\underline{0.056}   & 0.164   \\ \hline
Full             & \textbf{0.166}  & \textbf{0.191}   & \textbf{0.062}   & \textbf{0.173}   \\ \hline
\end{tabular}
}
\caption{
Ablation study on reminder task.
CHP stands for CoT-based hierarchical planning, Pos Prior represents the general area where obstacles are located in POLM, and POLM Prior is the pixel point where the filtered target is exactly located.
}
\vspace{-15pt}
\label{aba}
\end{table}
Table~\ref{aba} shows the ablation study of WalkVLM, to evaluate the effectiveness of CoT-based hierarchical planning (CHP) and POLM prior.
We conducted three sets of ablation experiments: 
\textbf{(a) w/o CHP.} Remove the CHP mechanism and generate reminders directly from the input visual information.
\textbf{(b) w/o Pos Prior.} Eliminates the approximate position of significant obstacles in POLM.
\textbf{(c) w/o POLM Prior.} Excludes the filtered target’s exact location and category as input. 
In these experiments, when the CHP mechanism was removed, the performance degradation was significant, which may be due to the model's inability to fully perceive the scene, resulting in the discrepancies between the generated reminder distribution and the GT distribution. 
The inclusion of CHP allows the model to conduct more detailed analysis from static attributes and scene summaries, thereby obtaining more concise and accurate results.
When removing POLM prior, the model's ROUGE performance is worse compared to lacking position prior, indicating that the model relies heavily on the visual details.


\vspace{-5pt}
\section{Conclusion}


This paper introduces WalkVLM, a vision-language model to offer walking guidance for visual-impaired individuals, along with a diverse and extensive benchmark for the task. By utilizing the chain of thought for hierarchical planning, WalkVLM is able to generate concise yet effective guidance. In addition, the usage of temporal-aware adaptive prediction further improves the efficiency to facilitate real-time processing. Comprehensive experiments demonstrated the usage of the benchmark and the effectiveness of WalkVLM over other VLMs.

{
    \small
    \bibliographystyle{ieeenat_fullname}
    \bibliography{main}
}

\clearpage

\onecolumn
\tableofcontents
\clearpage

\twocolumn
\section*{\centering Appendix}
\setcounter{section}{0}



\section{Walking Awareness Dataset}

\subsection{Data Regional Distribution}
\begin{figure*}
    \centering
    \includegraphics[width=1\linewidth]{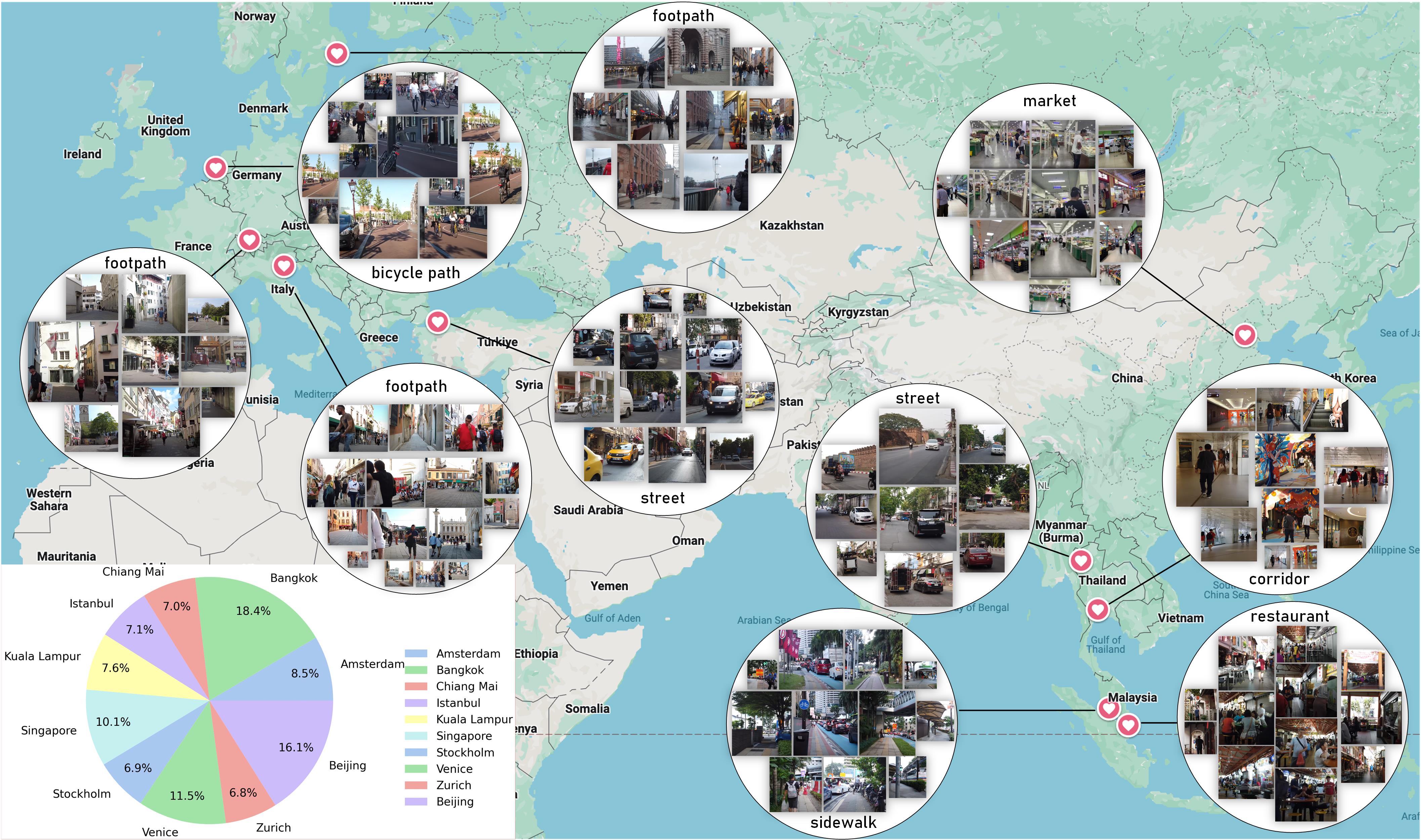}
    \caption{
    Visualization results of the WAD dataset sorted by region.
    The WAD dataset has a wide range of sources, and the samples and categories shown are randomly obtained from the dataset.
    The pie chart in the lower left corner shows the proportion of video length from different regions.
    }
    \label{fig:map}
\end{figure*}

Table \ref{data_q} shows the data distribution and corresponding duration in the WAD dataset.
The WAD dataset covers ten cities and contains a wide range of data sources.
Figure \ref{fig:map} illustrates the relevant regional distribution. 
As illustrated, our dataset is spread across Asia and Europe, showing a relatively balanced distribution between different regions.
Furthermore, the sampling across different regions is relatively uniform, with a large number of samples at various locations to avoid bias, which has good generalization characteristics.

\subsection{Dataset Category Definition}

As shown in Table \ref{categories}, WAD dataset contains multiple predefined data categories.
For weather conditions, we have selected the most common types, avoiding scenarios such as rainy or snowy days that make visually impaired people (VIPs) difficult to go outside.
For location types, we have selected the types where VIPs are likely to appear, avoiding rare locations.
For the traffic flow rating, we instructed annotators to count the number of people in each video segment and used this count as the basis for classification.
For scene summarization, during annotation, we required annotators to summarize static attributes such as road conditions, pedestrian flow, and vehicle flow, providing a comprehensive description of the current environment.
Currently, the granularity of our dataset is still relatively coarse. In the future, we will continue to refine different fine-grained categories and gradually expand the size of the dataset.

\begin{table}[h]
\centering
\begin{tabular}{lll}
\toprule
\textbf{City}         & \textbf{Country}   & \textbf{Hours}  \\ 
\midrule
Amsterdam    & Netherlands & 1:21h \\
Bangkok      & Thailand & 2:55h   \\
Chiang Mai   & Thailand & 1:07h   \\
Istanbul     & Turkey   & 1:08h   \\
Kuala Lampur & Malaysia & 1:12h   \\
Singapore    & Singapore & 1:36h   \\
Stockholm    & Sweden   & 1:06h   \\
Venice       & Italy    & 1:50h  \\
Zurich       & Switzerland & 1:05h \\ 
Beijing      & China  & 2:33h  \\
\bottomrule
\end{tabular}
\caption{The source region and duration of the WAD dataset. Refer to Fig. \ref{fig:map} for visualization results.}
\label{data_q}
\end{table}

\begin{table*}[]
\centering
\begin{tabular}{lll}
\hline
Tag Type                             & Category        & Note \\ \hline
\multirow{6}{*}{Weather Conditions}  & Sunny           &  -    \\
                                     & Night           &  Not make fine-grained distinctions     \\
                                     & Overcast        &  -    \\
                                     & Cloudy          &  -    \\
                                     & Indoor          &   Not make fine-grained distinctions   \\
                                     & Other           &   Severe weather conditions such as rain and fog for walking   \\ \hline
\multirow{8}{*}{Location Type}       & Busy Street     &  Open-air commercial streets    \\
                                     & Road            &    Roads where vehicles can travel normally  \\
                                     & Restaurant      &   Food stalls gathered together, inside large canteens   \\
                                     & Pedestrian Path &   Walking paths in parks and other places for healthy walking   \\
                                     & Corridor        &  Indoor walking paths    \\
                                     & Bicycle Lane    &   Bicycle roads with bicycle signs   \\
                                     & Shopping Mall   &    Large shopping supermarkets  \\
                                     & Other           &    Niche scenarios  \\ \hline
\multirow{3}{*}{Traffic Flow Rating} & Low             &   Fewer than 2 people appear in the sliced video   \\
                                     & Mid             &   Between 2 and 10 people appear in the sliced video   \\
                                     & High            &   More than 10 people appear in the sliced video   \\ \hline
\multirow{3}{*}{Danger Level}        & Low             &   The road is clear, the pedestrian flow is low, and no dangers within 15 steps   \\
                                     & Mid             &   Other scenarios that do not belong to low or high   \\
                                     & High            &     Potential collision factors, such as narrow roads, bumpy roads, vehicle warnings  \\ \hline
Scene Description & -                &  Detailed description of the current environment, level of danger, and pedestrian flow    \\ \hline
QA     & -                             &   The three types of inquiries mentioned in the paper and concise responses   \\
Reminder                    & -         &    Brief walking directions to provide to the user based on the current scenario  \\ \hline
\end{tabular}
\caption{The interpretation of label categories contained in the WAD dataset.}
\label{categories}
\end{table*}

\subsection{Annotation Process}
\begin{figure*}
    \centering
    \includegraphics[width=1\linewidth]{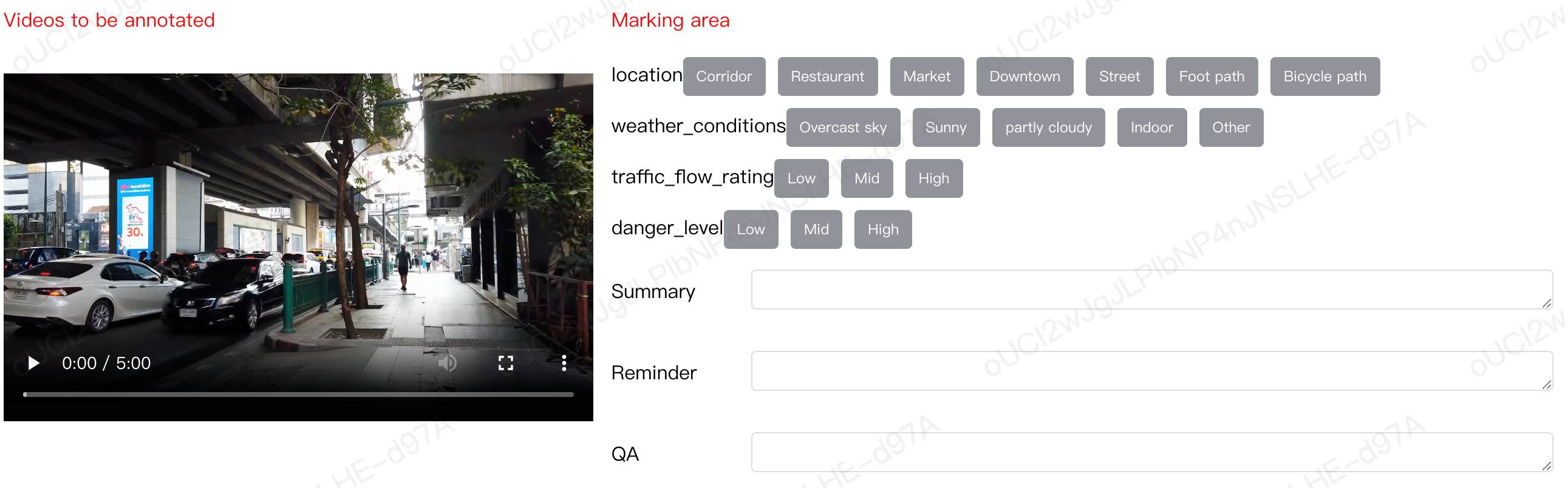}
    \caption{Annotation tool interface. Annotators mark the static attributes of the video in the video, record the time points of reminders and QA, and enter corresponding text descriptions.}
    \label{fig:annotation}
\end{figure*}

We use the page shown in Figure \ref{fig:annotation} to request annotators to make marks.
For static tags, we have provided relevant options for the annotators.
For scene summary, we require annotators to describe aspects such as the scene, road conditions, pedestrian flow, and vehicle flow.
For reminder and QA, we require annotators to expand on different situations, as described in Section \textcolor{red}{3.2} of the main paper.
Since descriptive tags carry a temporal dimension, we have adopted the annotation method in Table \ref{temporalann} for labeling.
After the text categorization is completed, we perform a quality inspection on it and use LLama3.1\footnote{https://ai.meta.com/blog/meta-llama-3-1/} to normalize the samples that pass the inspection to debias.

\begin{figure*}
    \centering
    \includegraphics[width=1\linewidth]{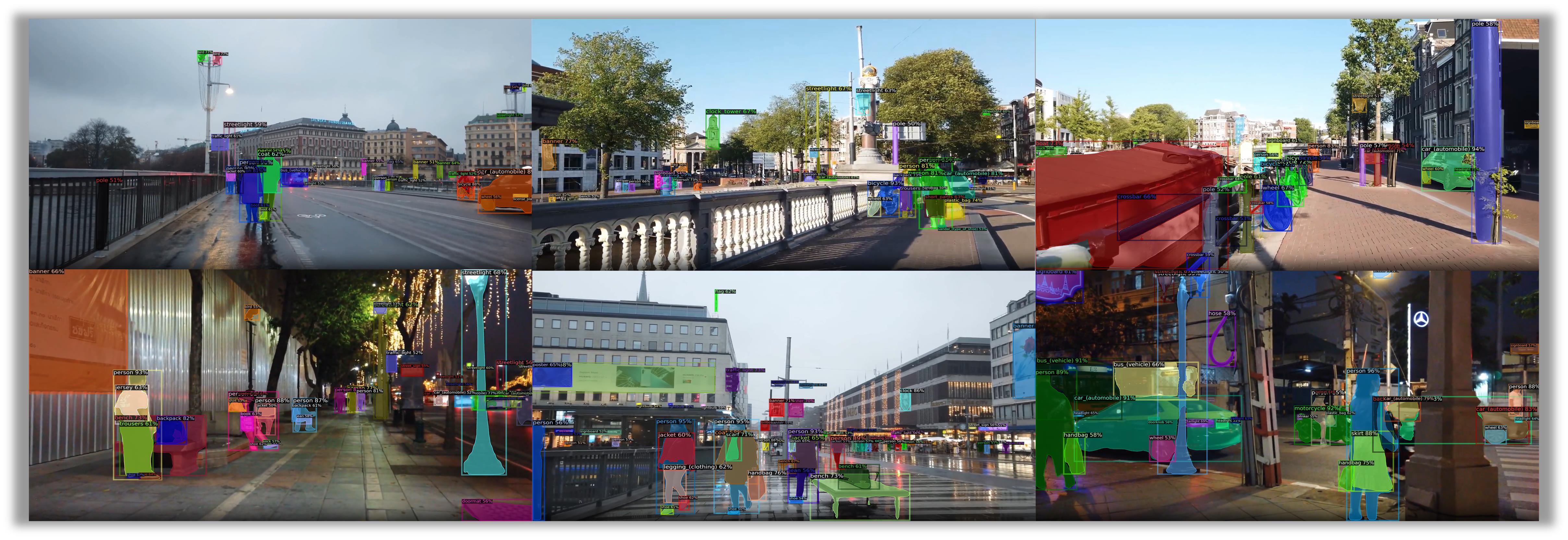}
    \caption{The detection results provided in the WAD dataset, which were pre-detected by the Detic model \cite{zhou2022detecting}, and then manually reviewed to ensure the correctness of the results.
    See \textcolor{blue}{\href{https://walkvlm2024.github.io}{here}} for more detection samples.
    }
    \label{fig:detectionresults}
\end{figure*}

\begin{figure*}
    \centering
    \includegraphics[width=0.9\linewidth]{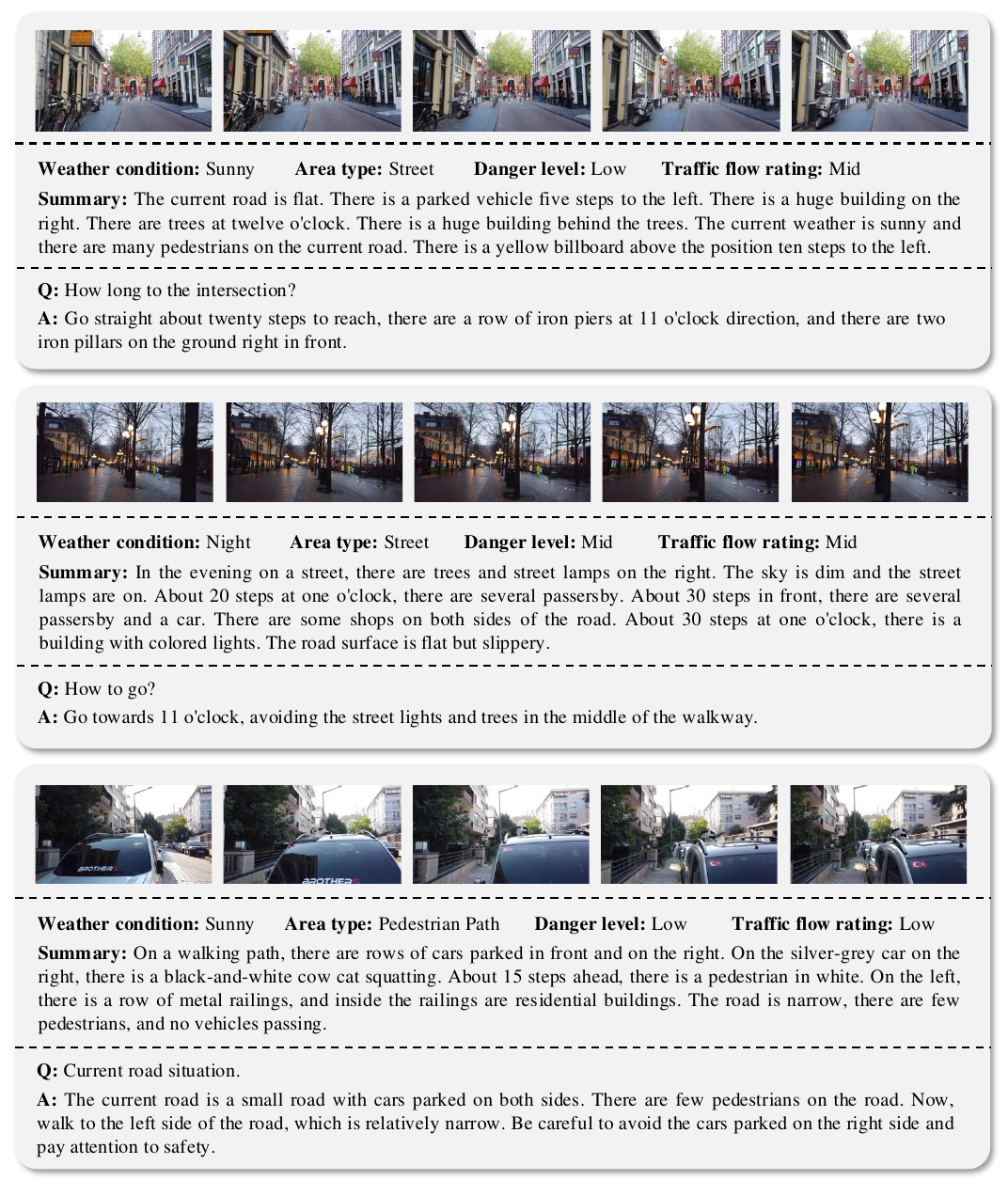}
    \caption{
    Visual examples of QA samples in WAD dataset.
    See \textcolor{blue}{\href{https://walkvlm2024.github.io}{here}} for dynamic samples.
    }
    \label{fig:qa_examples}
\end{figure*}

\begin{figure*}
    \centering
    \includegraphics[width=0.9\linewidth]{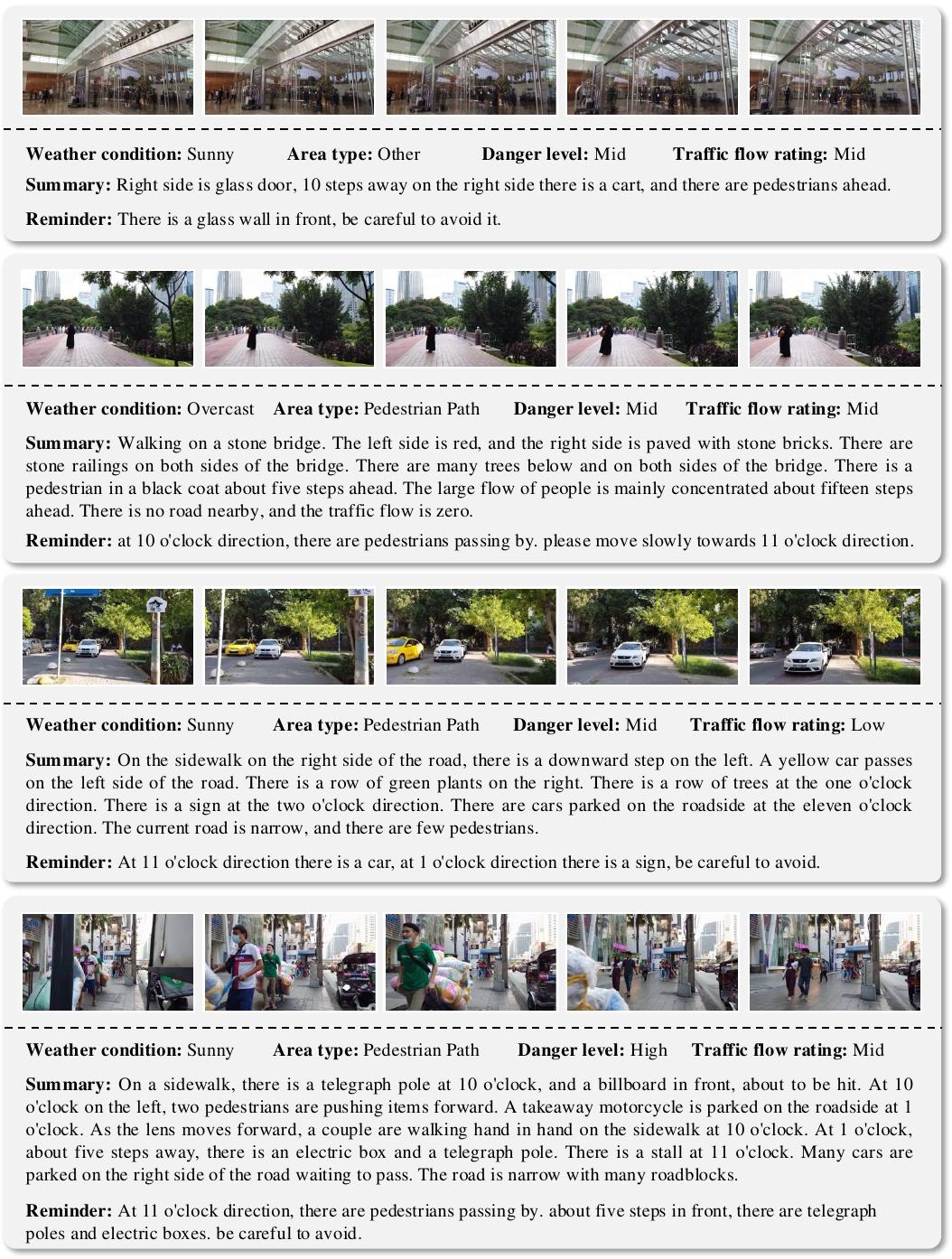}
    \caption{Visual examples of reminder samples in WAD dataset. See \textcolor{blue}{\href{https://walkvlm2024.github.io}{here}} for dynamic samples.
    }
    \label{fig:reminder_examples}
\end{figure*}

\begin{figure}
    \centering
    \includegraphics[width=1.0\linewidth]{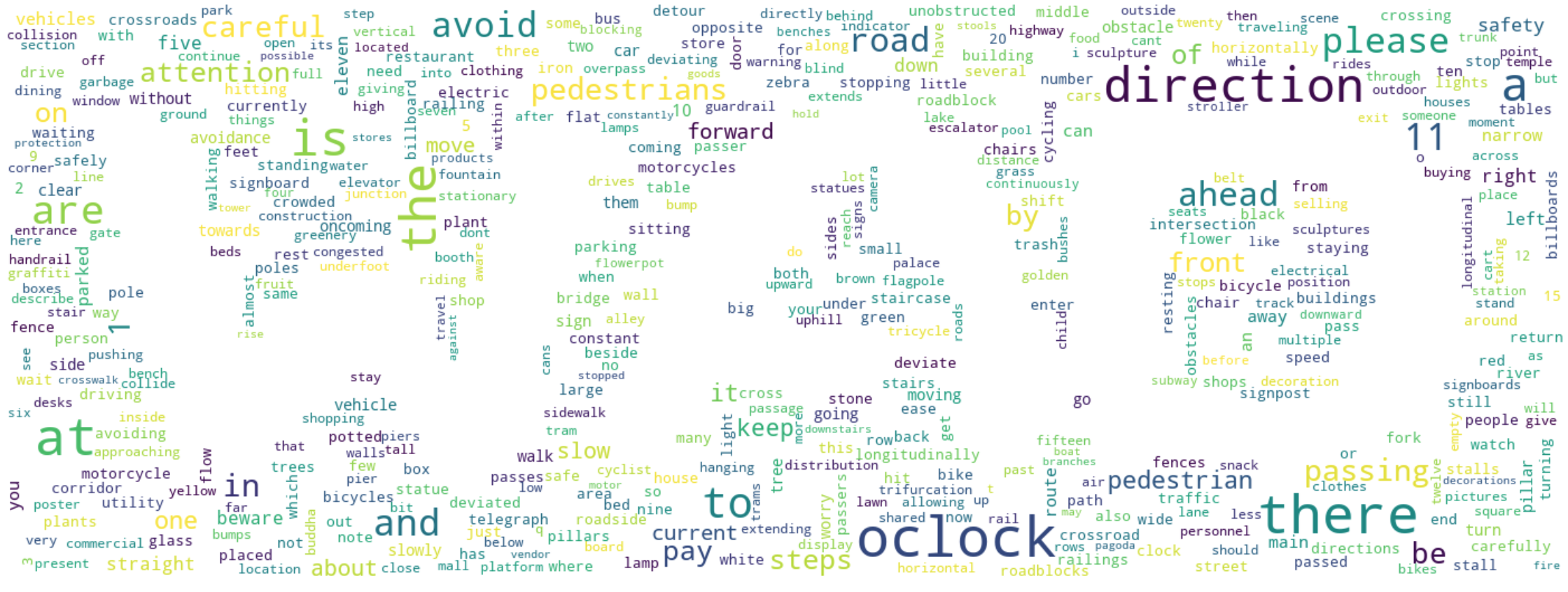}
    \caption{Word cloud distribution of the description in Walking Awareness Dataset. }
    \label{fig:cloud}
\end{figure}

\begin{figure*}
    \centering
    \includegraphics[width=1\linewidth]{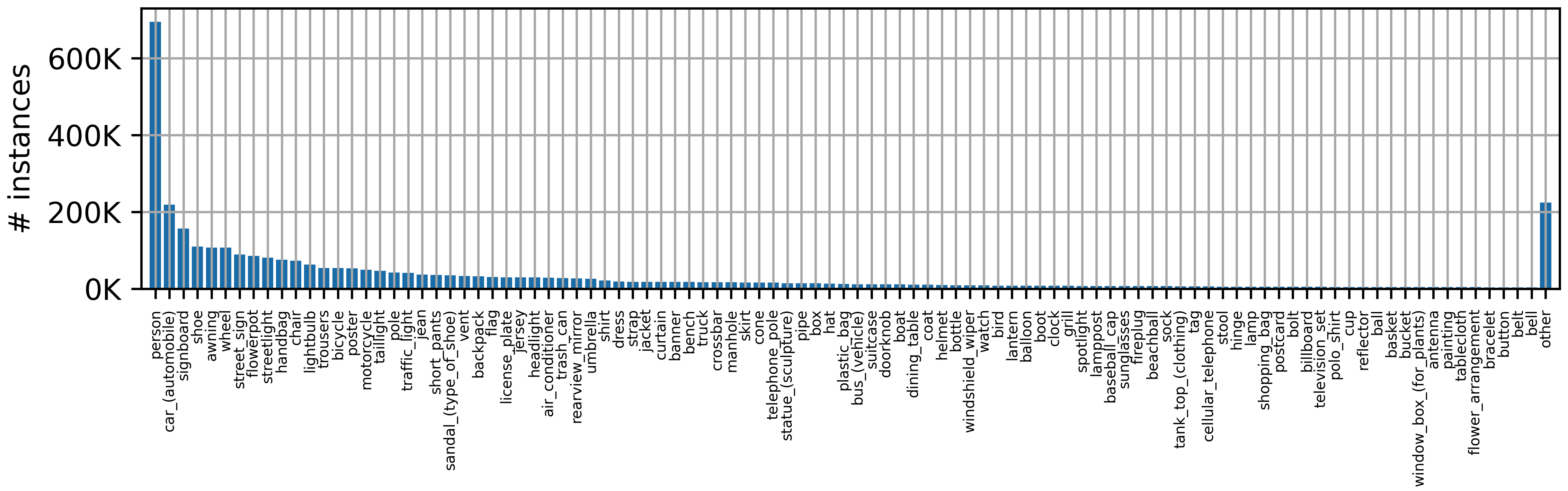}
    \caption{Detect target distribution. For clarity, display the top 100 with the highest frequency of occurrence.}
    \label{fig:target_detect}
\end{figure*}

\begin{figure*}
    \centering
    \includegraphics[width=1\linewidth]{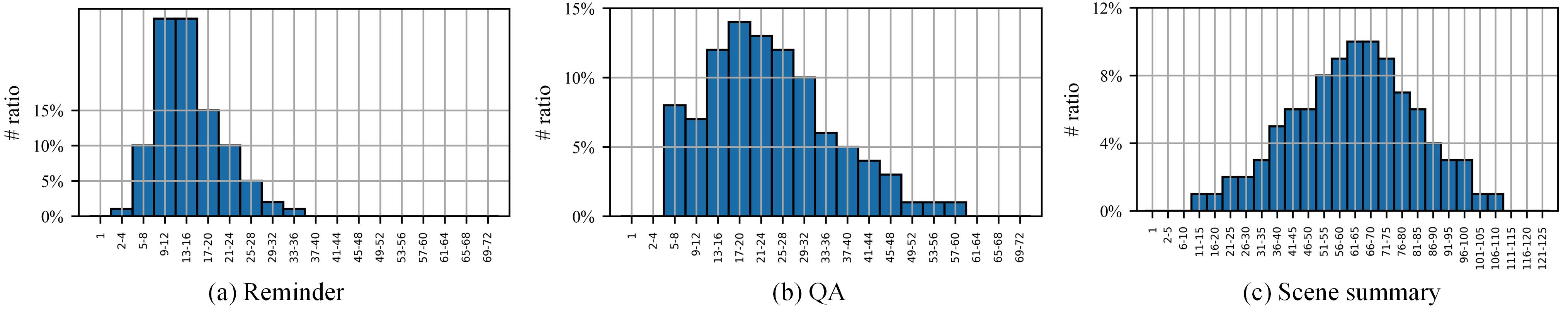}
    \caption{Data length distribution in different text annotation types.}
    \label{fig:instance}
    \vspace{-5px}
\end{figure*}

\begin{table}[]
\begin{tabular}{p{8cm}}

\toprule  

$\langle$ \textbf{time - category1,category2,...} $\rangle$

\textbf{Annotation} \\

\midrule

\textit{...}

$\langle$ 2m30s-A,E $\rangle$

\textit{almost hit the wall, go forward in the 11 o'clock direction to return to the main route.}

\textit{$\langle$ 2m43s-B $\rangle$}

\textit{five steps ahead is the fork in the road, go forward in the 10 - o'clock direction to return to the main route.}

\textit{...}

\textit{$\langle$ 3m39s-O $\rangle$}

\textit{Q: describe the current scene}

\textit{A: at a crossroads with many vehicles, keep still to avoid, there are some obstacles ahead, be careful to avoid}

\textit{...}
 \\ \bottomrule
\end{tabular}
\caption{Example of reminder and QA result annotation with a temporal dimension. We required annotators to mark the time when events occurred in the video, the question and reminder categories, as well as concise responses.}
\label{temporalann}
\end{table}

\subsection{Detection Model}

The Detic model \cite{zhou2022detecting} has achieved excellent results on the LVIS benchmark \cite{gupta2019lvis} in open-world detection tasks by training the detector classifier on image classification data.
In view of the model's good generalization ability, we use it to perform preliminary target extraction on the WAD dataset.
Figure \ref{fig:detectionresults} presents some example of the detection results of the Detic model in the WAD dataset, demonstrating that the model has a strong ability to extract small and complex targets.
After using the model for detection, we conducted manual confirmation and deleted some false positive boxes, thus obtaining the final detection results.

\begin{table*}[]
\begin{tabular}{p{17cm}}
\toprule  \scriptsize{\textit{person, car-(automobile), signboard, shoe, awning, wheel, street-sign, flowerpot, streetlight, handbag, chair, lightbulb, trousers, bicycle, poster, motorcycle, taillight, pole, traffic-light, jean, short-pants, sandal-(type-of-shoe), vent, backpack, flag, license-plate, jersey, headlight, air-conditioner, trash-can, rearview-mirror, umbrella, shirt, dress, strap, jacket, curtain, banner, bench, truck, crossbar, manhole, skirt, cone, telephone-pole, statue-(sculpture), pipe, box, hat, plastic-bag, bus-(vehicle), suitcase, doorknob, boat, dining-table, coat, helmet, bottle, windshield-wiper, watch, bird, lantern, balloon, boot, clock, grill, spotlight, lamppost, baseball-cap, sunglasses, fireplug, beachball, sock, tank-top-(clothing), tag, cellular-telephone, stool, hinge, lamp, shopping-bag, postcard, bolt, billboard, television-set, polo-shirt, cup, reflector, ball, basket, bucket, window-box-(for-plants), antenna, painting, tablecloth, flower-arrangement, bracelet, button, belt, bell, baby-buggy, flagpole, ladder, bowl, spectacles, vase, clock-tower, blouse, book, stop-sign, handle, banana, refrigerator, toy, sunhat, beanie, doughnut, necklace, train-(railroad-vehicle), bottle-cap, fan, tarp, vest, crate, orange-(fruit), magazine, apple, skateboard, parking-meter, postbox-(public), necktie, dog, earring, vending-machine, sweatshirt, barrel, lampshade, chandelier, cowboy-hat, minivan, newsstand, choker, hook, dish-antenna, scarf, camera, pizza, mask, drawer, weathervane, figurine, motor-scooter, magnet, pigeon, speaker-(stereo-equipment), cart, cooler-(for-food), blackboard, roller-skate, hot-air-balloon, flip-flop-(sandal), unicycle, headscarf, cabinet, hatbox, mirror, legging-(clothing), candle, satchel, teddy-bear, lanyard, log, glove, pennant, wall-socket, shower-cap, blinker, canister, pottery, robe, gargoyle, steering-wheel, newspaper, suspenders, dumpster, water-bottle, easel, kite, cushion, apron, horse, wreath, pew-(church-bench), dispenser, tomato, towel, melon, pumpkin, doormat, fire-extinguisher, sombrero, walking-cane, can, telephone-booth, thermostat, wineglass, heart, bandanna, tambourine, cat, jar, peach, carton, ring, frisbee, pot, carrot, watering-can, surfboard, mailbox-(at-home), headband, buoy, coconut, hose, card, sweater, lemon, remote-control, butterfly, grape, plate, knob, gravestone, knocker-(on-a-door), elephant, globe, mast, paper-plate, raincoat, wristlet, projector, watermelon, tote-bag, pirate-flag, mail-slot, tray, bulletproof-vest, brass-plaque, handcart, table, tricycle, towel-rack, laptop-computer, belt-buckle, fire-alarm, bow-(decorative-ribbons), slipper-(footwear), sink, papaya, sawhorse, briefcase, glass-(drink-container), cake, latch, coat-hanger, step-stool, fish-(food), napkin, pastry, motor, shopping-cart, sofa, silo, doll, toilet, tank-(storage-vessel), cookie, crucifix, oven, bamboo, tassel, hairnet, golfcart, fish, bread, cow, monitor-(computer-equipment) computer-monitor, lion, seashell, microwave-oven, earphone, Christmas-tree, water-jug, wagon-wheel, airplane, locker, broom, calendar, pop-(soda), barrette, mammoth, Rollerblade, avocado, blazer, scoreboard, hippopotamus, birdbath, shield, rubber-band, paper-towel, music-stool, straw-(for-drinking), poncho, neckerchief, pinwheel, houseboat, crutch, green-bean, birthday-card, sunflower, pickup-truck, grocery-bag, wine-bottle, faucet, halter-top, wine-bucket, sandwich, life-buoy, basketball-backboard, bullhorn, aerosol-can, tapestry, toilet-tissue, bathtub, tripod, goldfish, gourd, fireplace, stepladder, orange-juice, edible-corn, oil-lamp, garden-hose, potato, shower-curtain, water-tower, knife, onion, apricot, tennis-racket, piggy-bank, ashtray, puppet, sculpture, pretzel, fedora, brassiere, milk-can, cantaloup, blimp, blanket, guitar, kiwi-fruit, brake-light, armor, shawl, scissors, table-tennis-table, toothbrush, birdcage, lettuce, cylinder, radiator, turban, kimono, birdhouse, slide, envelope, Dixie-cup, Ferris-wheel, microphone, swimsuit, lime, beer-bottle, shaving-cream, fishbowl, ice-skate, camper-(vehicle), hairpin, pillow, underwear, oar, bonnet, chinaware, cymbal, penguin, sausage, strawberry, costume, dishtowel, gull, sword, bagel, spoon, crown, harmonium, duffel-bag, candle-holder, camcorder, horse-buggy, jumpsuit, clothes-hamper, knee-pad, bathrobe, comic-book, beer-can, giant-panda, map, phonograph-record, bell-pepper, toolbox, solar-array, rhinoceros, booklet, cupcake, shower-head, binoculars, monkey, matchbox, hand-towel, deer, pan-(for-cooking), dove, wheelchair, armoire, camel, goose, hair-dryer, dress-hat, tiger, tennis-ball, place-mat, bridal-gown, ottoman, cornice, mug, pear, sail, boxing-glove, passenger-car-(part-of-a-train), cap-(headwear), horse-carriage, urn, wig, wind-chime, thermos-bottle, fume-hood, crock-pot, bubble-gum, cherry, drum-(musical-instrument), wagon, bed, clarinet, eyepatch, tissue-paper, padlock, cigarette, parasol, baseball-bat, teacup, mandarin-orange, aquarium, bun, bowling-ball, telephone, lemonade, dog-collar, windmill, saltshaker, tartan, zucchini, lab-coat, tinsel, radar, pitcher-(vessel-for-liquid), pug-dog, sheep, coffee-maker, folding-chair, pinecone, visor, octopus-(animal), medicine, cassette, yogurt, saddlebag, wardrobe, basketball, persimmon, tape-(sticky-cloth-or-paper), tights-(clothing), baseball-glove, water-heater, cauliflower, cover, garbage-truck, forklift, bath-mat, chopping-board, computer-keyboard, propeller, wristband, gift-wrap, duck, railcar-(part-of-a-train), violin, football-helmet, blueberry, chopstick, piano, starfish, lawn-mower, fork, diaper, frying-pan, shark, wallet, duct-tape, pineapple, elk, toaster, earplug, wall-clock, cab-(taxi), zebra, bow-tie, hog, mallet, boiled-egg, knitting-needle, keycard, condiment, dragonfly, garlic, pepper-mill, drumstick, snowman, thumbtack, gasmask, pouch, teapot, sling-(bandage), barrow, bulldozer, spear, bookmark, mat-(gym-equipment), coffee-table, sleeping-bag, bat-(animal), runner-(carpet), iron-(for-clothing), bath-towel, coatrack, musical-instrument, bulletin-board, pie, tinfoil, overalls-(clothing), bib, pelican, egg, mascot, cistern, bookcase, giraffe, pad, trench-coat, bandage, chalice, flannel, clipboard, dustpan, celery, sweet-potato, headset, bread-bin, bowler-hat, walking-stick, saddle-blanket, phonebook, seahorse, clasp, lollipop, desk, broccoli, nailfile, anklet, dress-suit, rag-doll, beanbag, gondola-(boat), bear, mushroom, cider, dishwasher, alcohol, clementine, flap, rifle, ice-cream, ski, snowboard, vacuum-cleaner, automatic-washer, trailer-truck, hamper, television-camera, cigar-box, tobacco-pipe, bouquet, candy-bar, ferry, bead, banjo, ladybug, pacifier, shovel, control, fishing-rod, cruise-ship, washbasin, whipped-cream, pen, goggles, pan-(metal-container), flipper-(footwear), cucumber, nightshirt, dolphin, water-cooler, cloak, mop, pendulum, canoe, artichoke, heater, hammock, water-gun, almond, paintbrush, shredder-(for-paper), pita-(bread), liquor, eggbeater, scale-(measuring-instrument), dresser, ski-boot, cigarette-case, teakettle, armband, frog, file-cabinet, tow-truck, squid-(food), mouse-(computer-equipment), keg, tongs, deadbolt, quesadilla, hair-curler, koala, asparagus, platter, bobbin, coaster, milk, inhaler, salami, flamingo, life-jacket, coffeepot, urinal, eggplant, business-card, mattress, fig-(fruit), corkboard, raft, cash-register, cabana, suit-(clothing), kitchen-table, corset, gorilla, cocoa-(beverage), yacht, salmon-(fish), spice-rack, parachute, coil, squirrel, ironing-board, projectile-(weapon), coverall, trophy-cup, thread, measuring-stick, dinghy, crowbar, ski-pole, trunk, salad, dartboard, bedpan, award, rabbit, cincture, parka, colander, windsock, home-plate-(baseball), baboon, green-onion, eclair, toothpaste, saucer, highchair, handkerchief, pajamas, saxophone, potholder, ladle, spatula, first-aid-kit, veil, parakeet, scrubbing-brush, clip, blender, stapler-(stapling-machine), parrot, measuring-cup, owl, ice-maker, sweat-pants, videotape, corkscrew, marker, muffin, tiara, cast, beret, gun, tape-measure, generator, cowbell, sushi, hookah, seabird, crow, tachometer, cream-pitcher, battery, alligator, spider, Band-Aid, lightning-rod, hamburger, elevator-car, checkbook, hockey-stick, syringe, beeper, gelatin, wrench, water-scooter, hornet, fire-hose, Lego, stove, key, palette, chicken-(animal), deck-chair, chaise-longue, hairbrush, flashlight, smoothie, mitten, flute-glass, crab-(animal), bagpipe, clothespin, soap, lizard, river-boat, boom-microphone, radish, paperweight, fire-engine, candy-cane, bow-(weapon), sponge, wedding-cake, hourglass, ice-pack, tea-bag, cappuccino, eagle, machine-gun, salmon-(food), wet-suit, clutch-bag, cube, brussels-sprouts, wolf, toothpick, kennel, soccer-ball, prawn, hamster, identity-card, egg-yolk, pegboard, honey, duckling, pencil, ham, saddle-(on-an-animal), gameboard, hot-sauce, amplifier, alarm-clock, tortilla, manatee, brownie, nutcracker, popsicle, funnel, hotplate, trampoline, crib, heron, shampoo, butter, army-tank, date-(fruit), bottle-opener, cornet, camera-lens, jelly-bean, griddle, atomizer, armchair, bass-horn, hummingbird, salsa, baguet, sweatband, arctic-(type-of-shoe), footstool, power-shovel, drone, tractor-(farm-equipment), bunk-bed, food-processor, radio-receiver, cufflink, scarecrow, cock, cougar, chocolate-cake, wok, raspberry, ping-pong-ball, blackberry, dollhouse, space-shuttle, skewer, bobby-pin, school-bus, puffin, car-battery, razorblade, stirrup, drill, truffle-(chocolate), fighter-jet, thermometer, cupboard, screwdriver, sled, eel, pipe-bowl, broach, plume, sofa-bed, ferret, turtle, escargot, crescent-roll, printer, quilt, chocolate-bar, paddle, toaster-oven, motor-vehicle, puffer-(fish), soya-milk, cork-(bottle-plug), cabin-car, walrus, patty-(food), police-cruiser, skullcap, baseball, handsaw, Sharpie, stagecoach, cape, receipt, notebook, rib-(food), paperback-book, perfume, ballet-skirt, stirrer, steak-(food), telephoto-lens, barbell, record-player, mound-(baseball), dental-floss, sparkler-(fireworks), microscope, strainer, wooden-leg, dish, peeler-(tool-for-fruit-and-vegetables), hammer, milkshake, detergent, octopus-(food), limousine, chessboard, Tabasco-sauce, curling-iron, convertible-(automobile), underdrawers, freight-car, dalmatian, notepad, seaplane, burrito, dishrag, packet, birthday-cake, binder, wooden-spoon, pool-table, sewing-machine, pitchfork, cardigan, crayon, manger, kettle, CD-player, barge, flash, rolling-pin, cleansing-agent, dagger, waffle, hardback-book, toast-(food), puppy, egg-roll, chili-(vegetable), kitchen-sink, chocolate-mousse, router-(computer-equipment), pencil-sharpener, pin-(non-jewelry), kayak, sharpener, grater, nut, shoulder-bag, pantyhose, plow-(farm-equipment), mint-candy, crisp-(potato-chip), needle, pea-(food), beef-(food), sherbert, pepper, iPod, bullet-train, polar-bear, headboard, volleyball, bulldog, crape, reamer-(juicer), birdfeeder, table-lamp, pocketknife, jewelry, meatball, pudding, hand-glass, Bible, money, stylus, sugarcane-(plant), cayenne-(spice), shepherd-dog, lip-balm, soup-bowl, cornbread}}
 \\ \bottomrule
\end{tabular}
\caption{Full list of the target categories present in the walking awareness dataset, sorted by the number of occurrences in the dataset.}
\label{detect_fulllist}
\end{table*}

\subsection{Sample Visualization}
Figure \ref{fig:qa_examples} and Figure \ref{fig:reminder_examples} show more sample visualization results in the WAD dataset.
Our dataset has wide coverage, diverse types, and possesses ideal reminder attributes to train VLM to have guiding capabilities in blind walking tasks.

\subsection{Data Analysis}

Figure \ref{fig:target_detect} shows the distribution of the top 100 categories contained in the WAD dataset, while Table \ref{detect_fulllist} shows all the categories included.
Figure \ref{fig:cloud} presents a word cloud distribution with annotated descriptions, where the most frequently used words include \textit{oclock}, \textit{pedestrain}, \textit{direction}.
We have counted the word count distribution in different annotated texts in Figure \ref{fig:instance}. 
For reminder and QA scenarios, the data contained in WAD is shorter in length, while for summary scenario descriptions are more detailed.

\subsection{Benchmark Data Splits}

To ensure the diversity of test data, we adopted a category-based combined clustering method. Through this method, we carefully selected a certain number of samples from the clustering results to form our test set. Ultimately, we selected 1007 reminders and 134 QA pairs as our testset.
Furthermore, we conducted a thorough analysis of the distribution of the test set to confirm that they are accurate and that the same type of data is represented in the training set.

\subsection{Possible Sources of Bias}

Although the WAD dataset is collected from a wide range of geographical sources, we are aware of a few biases in our dataset.
The regions are still limited, which is still a long way from complete coverage of the globe.
The position of the camera and the divergence of focal length are also concerns for us, which need to obtain more general data to compensate for this.
In addition, the linguistic preferences of the annotators can introduce specific biases into the generated reminder, which implies that during the walking process, the model might provide information that are more appropriate for the area where the annotation was made.

\section{Model \& Details}

\subsection{All Prompts Used in Paper}

Table \ref{allprompts} displays all the prompts utilized in this paper under various circumstances such as normalizing annotation results, reasoning with VLM, and conducting evaluations.
Normalize the annotation results are crucial for ensuring the consistency and uniformity of annotation results, and this prompt are used in the preprocessing stage to correct bias in the data.
For the inference prompt of other models, we input historical multi-frame images and historical states to enable it to generate trigger states and reminders for the user.
In the prompt of WalkVLM, we make the model predict different levels of labels step by step and gradually output the results.
The evaluation prompt based on GPT4 compares different results with the ground truth to obtain the proportion statistics of the optimal model.

\subsection{Evaluation of Temporal Redundancy F1-Score}

This section systematically evaluates the redundancy of temporal outputs of different models.
Temporal redundancy refers to the excessive frequency of output information in this paper.
In order to evaluate the temporal redundancy of different models, we decompose the test video to ensure that each sample contains historical $N$ frames and $N$ states, thereby predicting the trigger state under the current situation.
We collected 834 such samples as a test set.
The predicted labels are divided into three levels, corresponding to the degree of danger. 
When the degree of danger is high, we regard it as triggering VLM.
By comparing the predicted different states with the ground truth, the distribution gap between the two sets of data can be calculated, thereby calculating the F1-score.

\begin{table*}[]
\begin{tabular}{p{2.0cm}p{3.3cm}p{11cm}}
\toprule
\centering \textbf{Application Scenario} & \centering \textbf{Models} & \textbf{Input Prompt} \\ \hline
    \textcolor{white}{a} 
    
    \textcolor{white}{a} 
    
 \centering  \small{  Normalize the annotation results }       &     \textcolor{white}{a}

    \textcolor{white}{a} 
    
  \centering  \small{ Llama 3.1}     &   \scriptsize{\textit{Please normalize the following manually annotated output to reduce information redundancy and maintain as standardized an output as possible. During the processing, please follow these guidelines: } 

\textit{a. Convert all labels to lowercase. }

\textit{b. Remove any superfluous spaces or special characters. }

\textit{c. Retain the accurate position of objects in the sentence, such as what time or how many steps away. }

\textit{d. For similar or repetitive semantic annotations, reduce the redundancy of semantics. }

\textit{e. Output the result directly. }

\noindent
\textit{The annotated text is as follows:} } 

...   \\ \hline

          \textcolor{white}{a} 
          
  \centering   \small{
       Instruct VLM to provide guidance on blind walking based on the provided video   }       &  
       
           \textcolor{white}{a} 
        
\centering  \small{
        DeepSeek (1.3B\&7B)

        Yi-VL (6B)

       MiniCPM-V2.6 (8B)

       GPT-4o
       
       Qwen2-VL (7B) }
       
       &   

\scriptsize{"request": "format the sentence below into the format, given in English",

"restriction": "answer in json like the format given below without code block",

"sentence": "Analyze the following video frames and determine the danger level for a blind person if they were to walk straight ahead. The danger levels are categorized as follows:\textbackslash{n}\textbackslash{n}A: Low (open areas, with few people or obstacles)\textbackslash{n}B: Mid (moderate danger, such as areas with some obstacles or moderate activity)\textbackslash{n} C: High (high danger, such as narrow pathways, crowded areas, or busy roads)\textbackslash{n}\textbackslash{n} Here are the danger levels for the first two frames:\textbackslash{n} Frame 1:\{history\_states[0]\}\textbackslash{n} Frame 2:\{history\_states[1]\}\textbackslash{n}\textbackslash{n} Based on this information and the provided image below, please provide the danger level for Frame 3.\textbackslash{n} Note that danger level are indicated by single letters only. (A, B, or C).\textbackslash{n} Additionally, provide walking instructions based on the provided image to ensure safe navigation.",

    "format": \{
        "data": \{
            "Frame 3 Danger Level": "string (A, B, or C)",
            "Walking Guidance": "string"
        \}
    \}}

       ...\\ \hline

    \textcolor{white}{a} 
    
    \textcolor{white}{a} 

        \textcolor{white}{a} 

\centering  \small{Training and inference of WalkVLM}

&            \textcolor{white}{a} 
        
    \textcolor{white}{a} 
    
    \textcolor{white}{a} 

        \textcolor{white}{a}

  \centering   \small{WalkVLM}
    &             \scriptsize{"request": "format the sentence below into the format, given in English",

            "restriction": "answer in json like the format given below without code block",

            "sentence": "You are now a guide. I can't see the path and will be walking solely based on your instructions. Each input frame displays the road information ahead. The main objects in each image are \{ json\_str \}. Please provide clear and unobstructed walking directions. Describe in order: 1. Location (e.g., corridor, restaurant, market, downtown, street, foot path, bicycle path), 2. Weather conditions (e.g., overcast sky, sunny, partly cloudy, indoor), 3. Traffic flow rating (e.g., low: 0-4 people/minute, medium: 4-10 people/minute, high: 10+ people/minute), 4. Describe the overall scene based on the input images and all the information from the above three points, 5. Please guide me on how to proceed based on the input images and all previous descriptions.",
            
            "format": \{
                "data": \{
                    "1. Location": "string",
                    "2. Weather conditions": "string",
                    "3. Traffic flow rating": "string",
                    "4. Describe the overall scene in the image": "string",
                    "5. Instructions on how I should proceed": "string"
                \}
            \}

            ...
            }

       \\ \hline

    \textcolor{white}{a} 
    
           \textcolor{white}{a} 
        
    \textcolor{white}{a} 
    
    \textcolor{white}{a} 
   
    \textcolor{white}{a} 
       
   \centering  \small{Use LMM to evaluate the similarity between generated results and ground truth}   &     

    \textcolor{white}{a} 
        
    \textcolor{white}{a} 
        
    \textcolor{white}{a} 
    
    \textcolor{white}{a} 
   
    \textcolor{white}{a} 
        
    \textcolor{white}{a}

    \textcolor{white}{a}

    \textcolor{white}{a} 
    
  \centering   \small{GPT4} &   \scriptsize{\textit{Please act as an impartial judge and evaluate the quality of the responses provided by multiple assistants displayed below. You should choose the assistant that matches the GT answer. Your evaluation should consider factors such as the helpfulness, relevance, accuracy, depth, creativity, and level of detail of their responses. Avoid any positional biases and ensure that the order in which the responses were presented does not influence your decision. Do not favor certain names of the assistants. Be as objective as possible. The answer should be the most closest to the semantics of the GT result and have the most concise answer. After providing your explanation, strictly follow the following format to output your final verdict: if assistant A is better, output "[[A]]", if assistant B is better, output "[[B]]". Request you select a relatively optimal result and directly output the option.}}

\textcolor{white}{a} 
    
\textit{\{GT\}}

\textit{\{\}}

\textcolor{white}{a} 
    
\textit{The Start of Assistant A's Answer}

\textit{\{\}}

\textit{The End of Assistant A's Answer}

\textcolor{white}{a} 

\textit{The Start of Assistant B's Answer}

\textit{\{\}}

\textit{The End of Assistant B's Answer}
    
% ...




       \\
       
\bottomrule
\end{tabular}
\caption{All prompts utilized in this paper.}
\label{allprompts}
\end{table*}

\section{Experiment}

\subsection{Visualization of Hierarchical Reasoning}

\begin{figure*}
    \centering
    \includegraphics[width=1\linewidth]{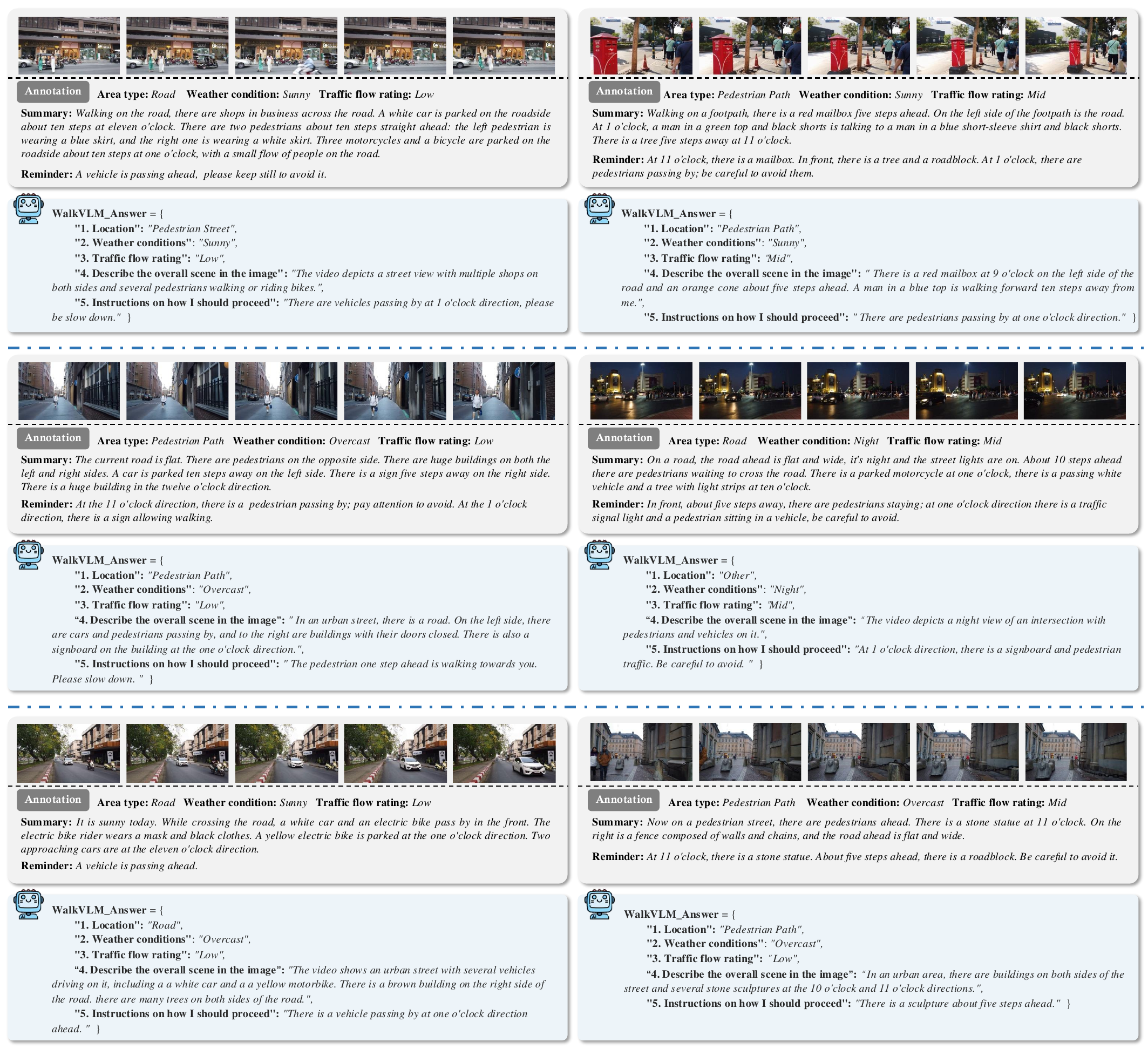}
    \caption{ Hierarchical inference results visualization of WalkVLM.}
\label{fig:hierarchical_visual}
\end{figure*}

We have demonstrated the results of hierarchical reasoning using WalkVLM in Figure \ref{fig:hierarchical_visual}.
WalkVLM can effectively extract static attributes from video streams and generate a comprehensive summary of the current scene.
After integrating fragmented attributes, the model produces concise and informative walking instructions.

\subsection{Visual Comparison of Different Models}

\begin{figure*}
    \centering
    \includegraphics[width=1\linewidth]{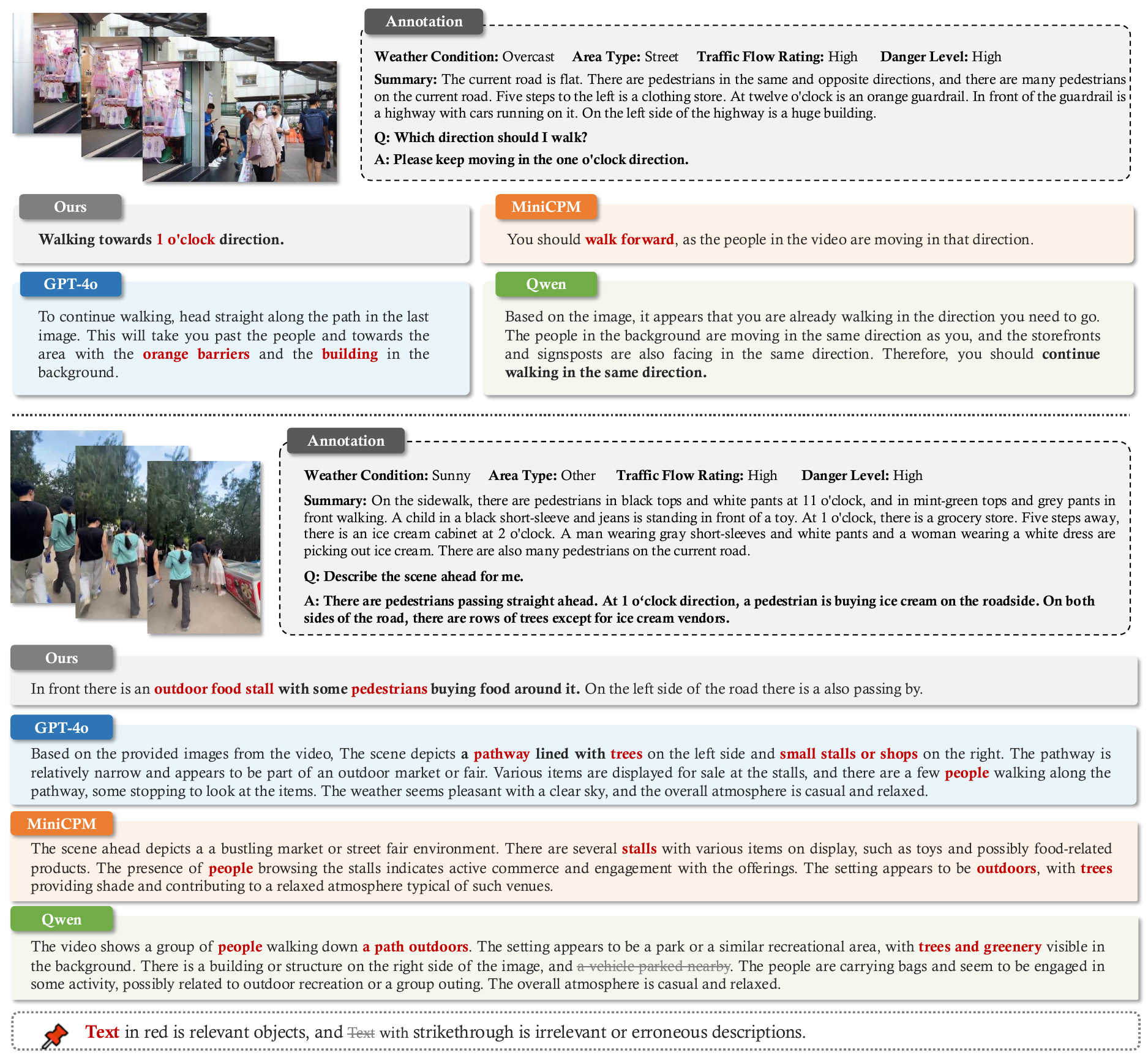}
    \caption{More QA visualization comparisons of different models.}
    \label{fig:qa_visual}
    \vspace{-10px}
\end{figure*}

\begin{figure*}
    \centering
    \includegraphics[width=1\linewidth]{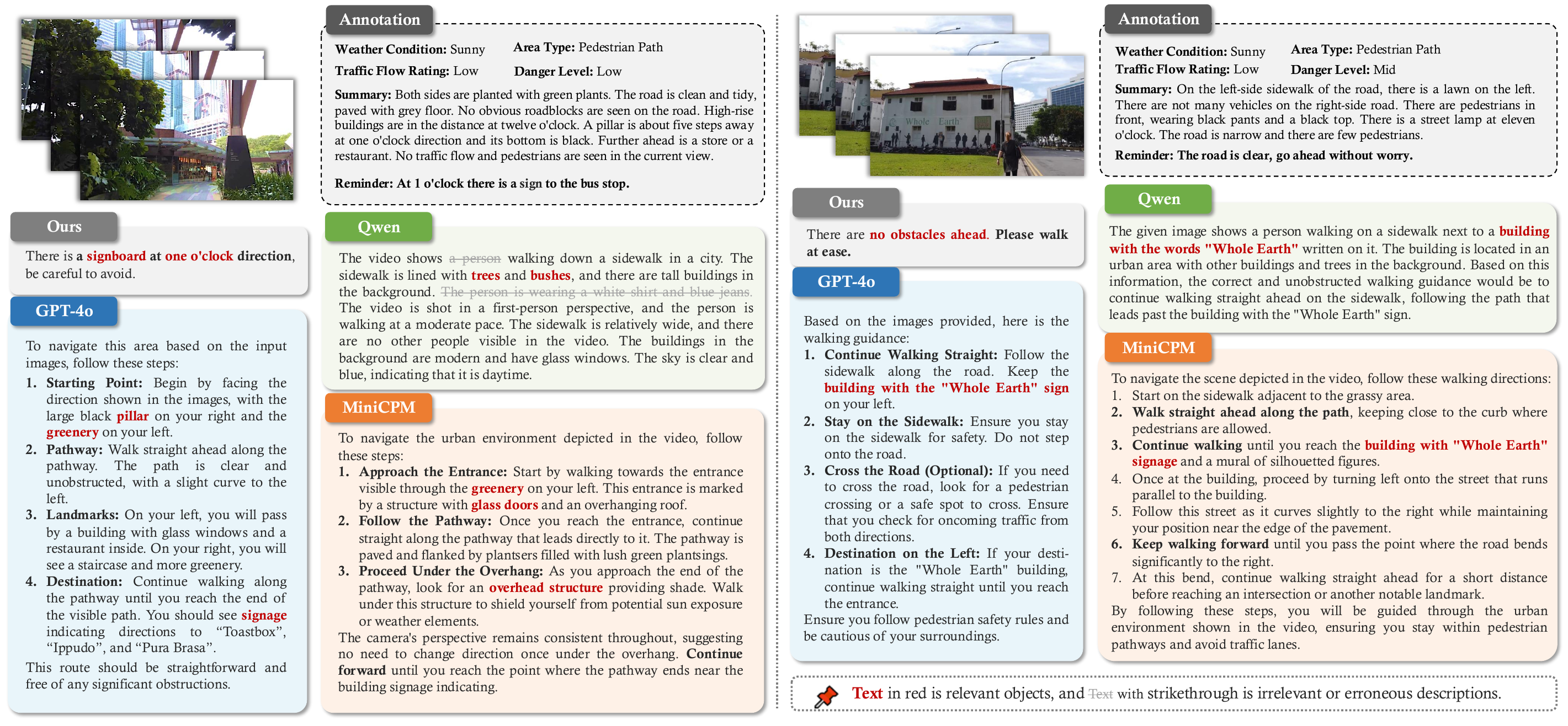}
    \caption{More reminder visualization comparisons of different models.}
    \label{fig:reminder_visual}
\end{figure*}

Figure \ref{fig:qa_visual} and \ref{fig:reminder_visual} presents a comparison of additional visualization results between WalkVLM and other models. Our approach yields more streamlined results, enabling a superior human-machine interaction experience during blind walking task.

\subsection{Comparison of Video Streaming Inference}

\begin{figure*}
    \centering
    \includegraphics[width=1\linewidth]{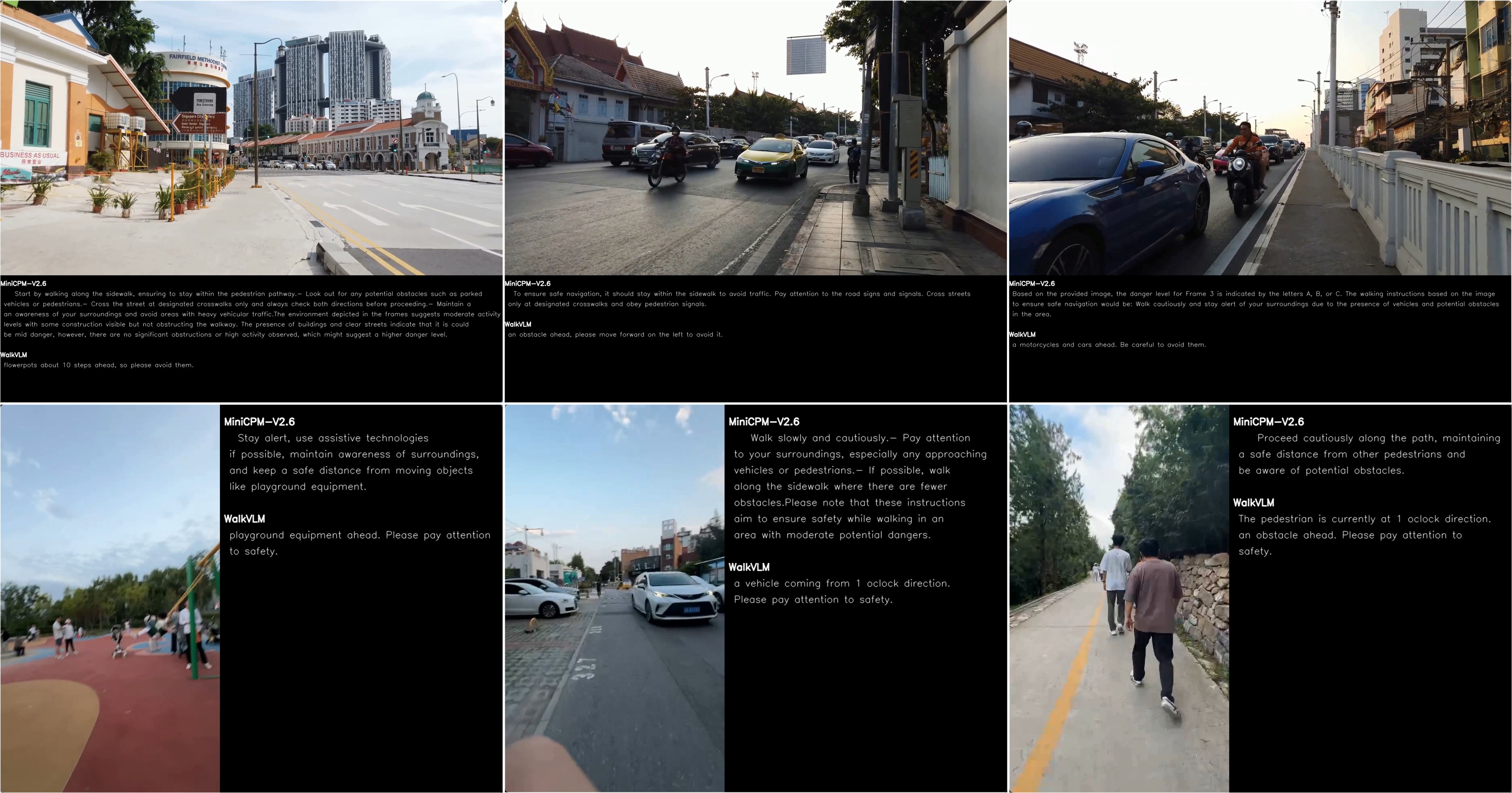}
    \caption{
Sampling results of video stream inference in the blind walking task.
Zoom in to view the generated results.
See \textcolor{blue}{\href{https://walkvlm2024.github.io}{here}} for the video inference results.
WalkVLM is capable of generating less temporal redundancy and providing more concise and informative responses.
    }
    \label{fig:video_stream}
\end{figure*}

In this section, we deployed WalkVLM and MiniCPM-V2.6 \cite{yao2024minicpm} on cloud devices to verify the differences in performance between the two models in real-world scenarios.
The visualization results of the two models on the video stream can be viewed \textcolor{blue}{\href{https://walkvlm2024.github.io}{here}}, where WalkVLM is capable of generating less temporal redundancy.
As shown in \ref{fig:video_stream}, on real-time video streams, for two models with the same size parameters, WalkVLM can generate more concise and accurate walking guidance.

However, the current model still has certain \textbf{limitations} in practical applications.
\textbf{Firstly}, the model has a weak ability to prioritize events, making it difficult to identify the most urgent actions that need reminders in the scene.
Facing this issue, our next attempt is to establish an event priority model that enables the model to propose necessary events and obtain priority results through ranking.
\textbf{Second}, the model still has certain misjudgments in obstacle recognition and direction.
Going forward, we will attempt to inject more prior knowledge about obstacles into the model and try to design some rule-based methods to verify the output of WalkVLM, so as to enhance its usability.
\textbf{Thirdly}, there is still significant room for improvement in the model's recognition of fine-grained obstacles.
We believe that this can be compensated for by collecting more available data.

Although there are the certain shortcomings, compared to other models, WalkVLM has made a solid advancement in the blind walking task. We will continue to iterate on this model to further enhance its usability in real-world scenarios!

\section{Discussion}

In the context of the increasingly popular vision-language model field, it is crucial to explore how to use it to address the daily challenges faced by visually impaired patients. Our work on the WalkVLM model and the walking awareness dataset represents a significant step in this direction, aiming to empower individuals with visual impairments through advanced technological solutions.

One of the most rewarding aspects of this research has been the opportunity to apply cutting-edge AI research to a problem that has profound real-world implications. We are deeply committed to leveraging technology to enhance the quality of life for everyone, and our work on WalkVLM exemplifies this mission. By providing a tool that can offer more accurate and context-aware guidance, we hope to make a tangible difference in the lives of blind individuals, enabling them to navigate their environments with greater independence and confidence.

However, we also recognize that our current approach has several limitations that need to be addressed to fully realize its potential. One major limitation is the geographical scope of our dataset, which currently covers only Europe and Asia. To develop a truly global solution, we need to expand our data collection efforts to include a wider range of regions and environments. This will ensure that our model can adapt to the diverse conditions and challenges faced by blind individuals around the world.

Another important consideration is the need for more real-time capabilities in our model. While WalkVLM offers significant advancements in understanding and interpreting walking-related data, achieving rapid inference is essential for practical applications. Real-time processing allows for immediate feedback and adjustments, which are critical for ensuring the safety and effectiveness of assistive technologies.

Additionally, integrating Retrieval-Augmented Generation (RAG) techniques \cite{lewis2020retrieval, gao2023retrieval} could further enhance the information provided by our model. By combining WalkVLM with RAG, we can incorporate a broader range of perspectives and data sources, leading to more informative and contextually relevant responses. This approach not only improves the accuracy and utility of our model but also fosters a more dynamic and interactive user experience.

In conclusion, while our work on WalkVLM achieves a significant advancement in the field of assistive technologies for the visually impaired, there is still much to be done. By addressing the limitations mentioned above and continuing to innovate, we hope to build on our current achievements and contribute to a future where technology empowers individuals with visual impairments to lead more independent and fulfilling lives. Our commitment to this cause remains unwavering, and we look forward to the next steps in this journey!

\section{Societal Impact}

Our contribution extends beyond the realm of technological advancement, offering significant societal benefits that can greatly improve the quality of life for visually impaired individuals. By introducing the WalkVLM model and the accompanying walking awareness dataset, we are taking a substantial step towards enhancing the independence and safety of blind individuals as they navigate through their daily environments.

Firstly, the WalkVLM model and dataset address a critical need for more accessible and effective assistive technologies for the visually impaired. Traditional navigation aids often fall short in providing the necessary real-time information and adaptability required for complex environments. Our model, with its advanced capabilities in understanding and interpreting walking-related data, can offer more precise and context-aware guidance, thereby reducing the risks associated with independent travel.

Moreover, the dataset we have compiled is a valuable resource that can foster further research and development in the field of assistive technologies. By making this dataset publicly available, we encourage collaboration and innovation among researchers, leading to the creation of even more sophisticated solutions that can cater to the diverse needs of blind individuals. This collaborative effort can ultimately result in technologies that are not only more effective but also more widely adopted and accessible.

From an educational standpoint, our work can also play a pivotal role in raising awareness about the challenges faced by the visually impaired community. By showcasing the potential of AI and machine learning in addressing these challenges, we hope to inspire more individuals and organizations to contribute towards creating a more inclusive society. This increased awareness can lead to more supportive policies and initiatives that focus on improving the quality of life for the visually impaired.

Additionally, the WalkVLM model and dataset have the potential to impact various industries beyond assistive technologies. For instance, they can be adapted for use in smart city planning, where understanding pedestrian behavior and safety is crucial. This broader application can lead to safer and more accessible urban environments for everyone, not just the visually impaired.

In summary, our contribution not only advances the state of the art in AI and machine learning but also has far-reaching societal implications. By providing a robust benchmark and a rich dataset, we are paving the way for innovative solutions that can significantly enhance the lives of blind individuals and promote a more inclusive society.

\section{Limitations}

This paper proposes a WAD dataset and systematically establishes the blind walking task based on the vision-language model, thereby setting up an extensive benchmark and offering valuable data support to this field.
Although the WAD dataset covers dozens of cities, its generalization capability is still relatively limited in practical applications, making the collection of additional data an essential endeavor.
Moreover, we devised the WalkVLM to make the reminders concise and opportune, but still leave considerable room in inference efficiency.



\end{document}